\documentclass[lettersize,journal]{IEEEtran}
\usepackage{amsmath,amsfonts}

\usepackage{algorithm}
\usepackage{array}
\usepackage[caption=false,font=normalsize,labelfont=sf,textfont=sf]{subfig}
\usepackage{textcomp}
\usepackage{stfloats}
\usepackage{url}
\usepackage{verbatim}
\usepackage{graphicx}
\usepackage{cite}

\usepackage{hyperref}

\usepackage{amsmath, amssymb, mathtools}

\usepackage{tabularx}
\usepackage{balance}

\usepackage{algpseudocode}
\hypersetup{hidelinks}
\usepackage{amsthm}
\usepackage{subcaption}
\usepackage{booktabs}

\usepackage{adjustbox}

\newcommand{\diffzeroone}{\textsf{\textsc{Diff-01}}}

\DeclareMathOperator*{\argmax}{arg\,max}

\begin{document}

\title{Expected Gain-based Escalation in Vertical Federated Learning}
\author{
\IEEEauthorblockN{
Mohamad Mestoukirdi,
Vincent Corlay
}

\IEEEauthorblockA{Mitsubishi Electric R\&D Centre, Rennes, France}

}
\maketitle

\begin{abstract}

Collaborative inference can improve predictive performance by integrating complementary information across agents, but applying collaborative fusion to every sample can incur unnecessary communication and computational overhead. 
This trade-off is particularly relevant in vertical federated learning (VFL), where clients observe different views of the same sample and fusion typically requires transmitting intermediate representations to a server. 
We study selective escalation in a two-round VFL inference protocol, in which a low-cost first round produces a prediction from client posteriors and a second embedding-fusion round is invoked only when it is expected to improve the final decision. 
We formulate routing as expected-gain score estimation: a sample is escalated when a predicted improvement in correctness justifies the additional communication. 
The proposed analytical score combines a calibrated pooled posterior with classwise reliability estimates of the VFL model, both obtained from held-out calibration data, yielding an interpretable router that requires no separately trained routing network. 
Experiments on multi-view classification benchmarks, including controlled test-time view degradation settings, show that the proposed router improves the communication--accuracy trade-off over confidence-, learned-gain-, and deferral-based baselines.

\end{abstract}
\section{Introduction}

Collaborative inference across multiple agents can improve prediction quality, but it also raises a practical question: when is collaboration actually worth its cost? In many multi-agent systems, different clients observe complementary views of the same sample, and a stronger prediction can be obtained by combining their information at a server. Vertical federated learning (VFL) provides a natural framework for this setting: each client processes its own local view and only selected information is communicated for joint prediction \cite{flpaper,khan2025vflreview}. The difficulty is that this more accurate collaborative route is also the most expensive one, as it may require transmitting intermediate representations rather than lightweight local predictions. This trade-off is especially relevant at inference time. A system may be able to form a quick prediction from local client outputs alone, while a stronger collaborative prediction is available if additional communication is allowed. In practice, it is rarely necessary to invoke that stronger route on every sample. Some inputs are already easy for the local route, while others are precisely the cases where escalation to collaborative inference is most useful. The challenge is therefore to decide, on a per-sample basis, whether the extra communication is justified.

We study this question in a two-round VFL protocol. In Round--1, each client sends a local posterior and the server forms a pooled prediction. In Round--2, if the sample is escalated, clients send higher-dimensional embeddings that are fused by a stronger collaborative model. The routing problem is to decide, after Round--1, whether switching to the Round--2 predictor is likely to improve correctness enough to justify the additional cost.

Our approach is based on a simple idea: treat escalation as an expected-gain decision. Rather than learning a separate router from routing labels, we estimate the utility of escalation analytically from quantities that can be computed on held-out calibration data. The resulting score depends on two ingredients: a calibrated version of the pooled Round--1 posterior, which estimates the plausible class labels after the first round, and classwise reliability statistics of the Round--2 VFL model, which capture how often the collaborative route is correct for each class. Together, these quantities yield a threshold-based routing rule that is easy to interpret and straightforward to deploy.

This work makes the following contributions:
\begin{itemize}
    \item We study inference-time selective escalation in a two-round collaborative inference setting, where a lightweight initial prediction route can optionally invoke a costly collaborative VFL model.
    \item We propose an analytical routing rule based on expected gain, using only the pooled Round--1 posterior and classwise reliability statistics of the collaborative model.
    \item We empirically validate the proposed approach across multiple multi-view classification tasks and deployment conditions, showing that it achieves a favorable accuracy--escalation trade-off and consistently outperforms learned-gain-, confidence-, and deferral-based baselines.
\end{itemize}

\section{Related Work}

Our work sits at the intersection of communication-efficient vertical federated learning, selective prediction and deferral, and adaptive inference.

A substantial body of work in VFL studies how to reduce the communication and computation cost of collaborative training and inference. Existing directions include compression and quantization, sparse communication, feature selection, limited-overlap protocols, and caching or reuse of intermediate representations \cite{khan2022commvfl,castiglia2023lessvfl,inoue2023sparsevfl,sun2023oneshotvfl,zhou2025vflcafe}. These approaches are primarily concerned with making collaborative inference cheaper once it is performed. By contrast, our focus is on a different question: given a trained collaborative route, when should it be invoked at all at test time?

Our work is also related to selective prediction and reject-option classification, where a model may abstain when it is uncertain \cite{franc2023reject,hendrickx2024survey}, and more closely to learning-to-defer methods, where a predictor decides whether to keep its own prediction or hand the example to another model or expert \cite{madras2018defer,mozannar2020consistent,mao2023twostage,liu2024underfitting}. These lines of work provide the broader decision-theoretic context for our setting: a weak route is available immediately, while a stronger alternative may be used selectively. 

The closest conceptual connection is to adaptive inference and classifier cascades, where additional computation is used only when its expected benefit justifies the extra cost \cite{gao2011active,jitkrittum2023cascade}. Our two-round protocol fits naturally into this perspective: Round--1 provides a cheap pooled prediction, and Round--2 provides a stronger collaborative one. The main difference is in how the routing signal is constructed. Rather than learning a separate router or explicitly modeling the unseen output of the stronger route, we propose a plug-in approximation to the expected gain of escalation based on two ingredients estimated on held-out data: a calibrated posterior from the pooled Round--1 prediction and classwise reliability statistics of the Round--2 model.

\section{Problem Setup}
\label{sec:problem_setup}

We consider a multi-agent classification problem with $M$ clients and $K$ classes in a vertical federated inference setting, where different clients observe complementary partial views of the same underlying sample and a server combines the communicated information to produce a prediction \cite{flpaper}. A sample is drawn from an unknown distribution over
\[
(\boldsymbol{x}_1,\ldots,\boldsymbol{x}_M,y)\in \mathcal{X}_1\times\cdots\times\mathcal{X}_M\times\{0,\ldots,K-1\},
\]
where $\boldsymbol{x}_i$ denotes the observation available to client $i$ and $y$ is the ground-truth class label.

Each client $i\in\{1,\ldots,M\}$ is equipped with:
\begin{itemize}
    \item a local classifier
    \[
    f_i:\mathcal{X}_i\to \Delta^{K-1},
    \]
    which outputs a local posterior
    \[
    \boldsymbol{p}_i = f_i(\boldsymbol{x}_i)\in \Delta^{K-1},
    \]
    where $\Delta^{K-1}$ denotes the probability simplex over $K$ classes;
    \item a feature encoder
    \[
    E_i:\mathcal{X}_i\to \mathbb{R}^{d_i},
    \]
    which outputs an embedding
    \[
    \boldsymbol{e}_i = E_i(\boldsymbol{x}_i).
    \]
\end{itemize}

\subsection{Two-Round Inference}
\label{subsec:two_round_inference}

We study a two-round inference protocol in which a cheap Round--1 local route prediction may optionally be replaced by a stronger but more communication-intensive collaborative prediction.

\paragraph{Round--1: local posterior communication.}
Each client sends only its local posterior $\boldsymbol{p}_i$ to the server. The server constructs an aggregate posterior by average pooling:
\begin{equation}
\bar{\boldsymbol p}
=
\frac{1}{M}\sum_{i=1}^M \boldsymbol p_i .
\label{eq:avg_pool}
\end{equation}
The corresponding cheap prediction route is
\begin{equation}
\hat y_{\mathrm{loc}}
=
\argmax_{k\in\{0,\ldots,K-1\}} \bar{\boldsymbol p}(k).
\label{eq:y_loc}
\end{equation}

\paragraph{Round--2: collaborative embedding fusion.}
If the sample is escalated, each client computes an embedding $\boldsymbol e_i=E_i(\boldsymbol x_i)$ and transmits it to the server. A server-side VFL fusion model
\[
g_{\mathrm{VFL}}:\mathbb{R}^{d_1}\times\cdots\times\mathbb{R}^{d_M}\to \Delta^{K-1}
\]
produces the collaborative posterior
\begin{equation}
\boldsymbol q = g_{\mathrm{VFL}}(\boldsymbol e_1,\ldots,\boldsymbol e_M),
\label{eq:vfl_posterior}
\end{equation}
with prediction
\begin{equation}
\hat y_{\mathrm{VFL}}
=
\argmax_{k\in\{0,\ldots,K-1\}} \boldsymbol q(k).
\label{eq:y_vfl}
\end{equation}

Round--1 is intentionally lightweight: it requires only communication of $K$-dimensional posterior vectors and uses low-complexity local predictors. Round--2 is generally more expressive but more expensive, because it requires transmitting higher-dimensional intermediate representations and applying a collaborative server-side fusion model.

\subsection{Routing Policy and Objective}
\label{subsec:routing_objective}

A routing policy decides, after observing the pooled Round--1 posterior, whether the sample should be escalated to Round--2. Let
\[
\pi:\Delta^{K-1}\to\{0,1\}
\]
be a binary routing rule, where $\pi(\bar{\boldsymbol p})=1$ means \emph{escalate} and $\pi(\bar{\boldsymbol p})=0$ means \emph{adopt the Round--1 prediction}.

Given $\pi$, the final prediction is
\begin{equation}
\hat y^{\pi}
=
\begin{cases}
\hat y_{\mathrm{VFL}}, & \text{if } \pi(\bar{\boldsymbol p})=1,\\
\hat y_{\mathrm{loc}}, & \text{otherwise.}
\end{cases}
\label{eq:final_prediction}
\end{equation}

The system-level goal is to maximize prediction accuracy under an average escalation budget:

\begin{equation}
\begin{aligned}
\max_{\pi} \quad
& \mathbb{E}_{(\boldsymbol{x}_1,\ldots,\boldsymbol{x}_M,y)\sim \mathcal P}
\left[
\mathbf{1}\!\left[\hat y^{\pi}=y\right]
\right] \\
\text{s.t.} \quad
& \mathbb{E}_{(\boldsymbol{x}_1,\ldots,\boldsymbol{x}_M,y)\sim \mathcal P}
\left[
\pi(\bar{\boldsymbol p})
\right]
\le B
\end{aligned}
\label{eq:system_objective}
\end{equation}
where $\mathcal P$ denotes the test-time distribution and $B\in[0,1]$ is a prescribed escalation budget. The constraint is equivalent to controlling the average communication cost up to a constant factor\footnote{While our focus is not on optimizing the communication layer itself through compression, quantization, or coding techniques, we explicitly study selective escalation of the collaborative route as a mechanism for reducing communication cost at inference time. A more detailed discussion of the communication cost in this work is provided in Appendix.}. 

\subsection{Escalation Gain}
\label{subsec:escalation_gain}

For a sample with true label $y$, we define the \emph{gain of escalation} as
\begin{equation}
G
=
\mathbf{1}\!\left[\hat y_{\mathrm{VFL}}=y\right]
-
\mathbf{1}\!\left[\hat y_{\mathrm{loc}}=y\right]
\in\{-1,0,+1\}.
\label{eq:gain}
\end{equation}
This quantity distinguishes three cases:
\begin{itemize}
    \item $G=+1$: escalation is beneficial (Round--1 is wrong, Round--2 is correct);
    \item $G=0$: escalation is neutral (both routes agree in correctness);
    \item $G=-1$: escalation is harmful (Round--1 is correct, Round--2 is wrong).
\end{itemize}

This formulation follows the general decision-theoretic view that selective escalation should depend on the conditional benefit of invoking a stronger but more expensive prediction route. Closely related ideas appear in value-of-information-based active classification and in recent analyses of classifier cascades, where the routing decision is expressed through the relative correctness of a weak and a strong predictor \cite{jitkrittum2023cascade}. In our setting, the weak route is the pooled Round--1 predictor and the strong route is the collaborative Round--2 VFL predictor.

Since the escalation decision is made after Round--1, the pooled posterior $\bar{\boldsymbol p}$ is the natural observable available to the router. We therefore consider threshold-based policies of the form
\begin{equation}
\mathbb{E}[G\mid \bar{\boldsymbol p}] > \tau,
\label{eq:decision_rule_general}
\end{equation}
where $\tau$ controls the communication--accuracy trade-off. Using the definition of $G$, the conditional expected gain can be written as
\begin{align}
\mathbb{E}[G\mid \bar{\boldsymbol p}]
&=
\mathbb{P}(\hat y_{\mathrm{VFL}}=y\mid \bar{\boldsymbol p})
-
\mathbb{P}(\hat y_{\mathrm{loc}}=y\mid \bar{\boldsymbol p}).
\label{eq:gain_decomposition}
\end{align}
\section{Method}
\subsection{A Plug-in Approximation of the Expected Escalation Gain}
\label{subsec:plugin_approximation}
Conditioning on $\bar{\boldsymbol p}$, the first term in \eqref{eq:gain_decomposition} can be expanded as
\begin{align}
\mathbb{P}(\hat y_{\mathrm{VFL}}=y \mid \bar{\boldsymbol p})
&=
\sum_{k=0}^{K-1}
\mathbb{P}(\hat y_{\mathrm{VFL}}=y, y=k \mid \bar{\boldsymbol p})
\nonumber\\
&=
\sum_{k=0}^{K-1}
\mathbb{P}(\hat y_{\mathrm{VFL}}=k \mid y=k,\bar{\boldsymbol p})\,
\mathbb{P}(y=k \mid \bar{\boldsymbol p}).
\label{eq:vfl_term_exact}
\end{align}

The quantity $\mathbb{P}(\hat y_{\mathrm{VFL}}=k \mid y=k,\bar{\boldsymbol p})$ is generally sample-dependent and not directly available at test time. We therefore adopt the following approximation.

\paragraph{Classwise reliability approximation.}
We approximate the conditional correctness of the Round--2 route given the class by a classwise average reliability term:
\begin{equation}
\mathbb{P}(\hat y_{\mathrm{VFL}}=k \mid y=k,\bar{\boldsymbol p})
\approx
\mathbb{P}(\hat y_{\mathrm{VFL}}=k \mid y=k)
=
C_{\mathrm{VFL}}[k,k],
\label{eq:classwise_reliability_assumption}
\end{equation}
where \(C_{\mathrm{VFL}}[k,k]\) denotes the diagonal entry of the Round--2 confusion matrix for class \(k\). In other words, once the true class is fixed, we assume that the remaining dependence of Round--2 correctness on \(\bar{\boldsymbol p}\) is small enough to ignore in the analytical model. This makes the score easy to estimate from held-out data: the behavior of the collaborative route is summarized by one reliability coefficient per class instead of a fully sample-dependent conditional model. The trade-off is that the approximation cannot capture within-class variation in Round--2 behavior. If the collaborative model behaves differently for samples from the same class depending on the pooled Round--1 posterior, then the analytical score may systematically overestimate or underestimate the true benefit of escalation. This issue can become more pronounced under distribution shift, when the calibration data are not representative of test-time conditions, or when some classes are too rare to estimate their reliability coefficients accurately. Under the  approximation in \eqref{eq:classwise_reliability_assumption}, the first term in \eqref{eq:gain_decomposition} becomes
\begin{equation}
\mathbb{P}(\hat y_{\mathrm{VFL}}=y \mid \bar{\boldsymbol p})
\approx
\sum_{k=0}^{K-1}
C_{\mathrm{VFL}}[k,k]\,
\mathbb{P}(y=k \mid \bar{\boldsymbol p}).
\label{eq:vfl_term_plugin}
\end{equation}

For the Round--1 route, \(\hat y_{\mathrm{loc}}\) is a deterministic function of \(\bar{\boldsymbol p}\). Therefore,
\[
\mathbb{P}(\hat y_{\mathrm{loc}}=k \mid y=k,\bar{\boldsymbol p})
=
\mathbf{1}[\hat y_{\mathrm{loc}}=k],
\]
and we obtain:
\begin{align}
\mathbb{P}(\hat y_{\mathrm{loc}}=y \mid \bar{\boldsymbol p})
&=
\sum_{k=0}^{K-1}
\mathbb{P}(\hat y_{\mathrm{loc}}=k \mid y=k,\bar{\boldsymbol p})\,
\mathbb{P}(y=k \mid \bar{\boldsymbol p})
\nonumber\\
&=
\sum_{k=0}^{K-1}
\mathbf{1}\!\left[\hat y_{\mathrm{loc}}=k\right]\,
\mathbb{P}(y=k \mid \bar{\boldsymbol p}).
\label{eq:loc_term_exact}
\end{align}

Substituting \eqref{eq:vfl_term_plugin} and \eqref{eq:loc_term_exact} into \eqref{eq:gain_decomposition} yields the plug-in approximation
\begin{equation}
s(\bar{\boldsymbol p})
:=
\sum_{k=0}^{K-1}
\mathbb{P}(y=k \mid \bar{\boldsymbol p})
\left(
C_{\mathrm{VFL}}[k,k]
-
\mathbf{1}\!\left[\hat y_{\mathrm{loc}}=k\right]
\right),
\label{eq:analytical_score}
\end{equation}
which we use as the analytical routing score. The score is high when the calibrated posterior assigns low probability to the Round--1 predicted class while assigning probability mass to classes on which the Round--2 model is reliable. It becomes negative when the pooled posterior is concentrated on the current Round--1 label or when the Round--2 model is weak on the plausible classes, indicating that escalation is unlikely to help.
\subsection{Estimating the Score from Held-Out Calibration Data}
\label{subsec:score_estimation}

To evaluate the analytical score in \eqref{eq:analytical_score} at test time, two quantities must be estimated from held-out calibration data. The first is the conditional class posterior
\(
\mathbb{P}(y=k \mid \bar{\boldsymbol p})
\),
which captures how the pooled Round--1 posterior relates to the true label. The second is the classwise reliability of the Round--2 collaborative predictor, summarized by the diagonal entries
\(
C_{\mathrm{VFL}}[k,k]
\)
of its confusion matrix. We estimate these two components separately and then combine them into a deployable routing score.

\paragraph{Estimating Round--2 classwise reliability.}

Let
\[
\mathcal{D}_{\mathrm{calib}}
=
\left\{(\boldsymbol{x}_1^{(n)},\ldots,\boldsymbol{x}_M^{(n)},y^{(n)})\right\}_{n=1}^{N_{\mathrm{calib}}}
\]
be a held-out calibration set. Using the Round--2 VFL predictor on this set, we estimate the confusion matrix by
\begin{equation}
C_{\mathrm{VFL}}[k,c]
=
\frac{
\sum_{n:\,y^{(n)}=k}
\mathbf{1}\!\left[\hat y_{\mathrm{VFL}}^{(n)}=c\right]
}{
\sum_{n:\,y^{(n)}=k} 1
},
\label{eq:cvfl_estimator}
\end{equation}
so that the diagonal term \(C_{\mathrm{VFL}}[k,k]\) estimates the empirical probability that the collaborative route predicts class \(k\) correctly conditional on \(y=k\).

\paragraph{Calibrating the pooled Round--1 posterior.}
The remaining ingredient is the conditional distribution
\(
\mathbb{P}(y=k \mid \bar{\boldsymbol p})
\).
A natural baseline is to use the pooled posterior entry \(\bar{\boldsymbol p}(k)\) itself as an estimate of \(\mathbb{P}(y=k \mid \bar{\boldsymbol p})\). In practice, however, pooled posteriors are not perfectly calibrated. We therefore apply a multiclass Dirichlet calibration map \cite{calibration_d}\footnote{We compare against temperature scaling \cite{pmlr-v70-guo17a} and a no-calibration setting in the Appendix. In our experiments, Dirichlet calibration yields the best routing performance.} fitted on the same held-out calibration data:
\begin{equation}
\hat{\boldsymbol m}_{\phi}(\bar{\boldsymbol p})
=
\mathrm{softmax}\!\left(
W\log(\bar{\boldsymbol p}+\boldsymbol \epsilon)+\boldsymbol b
\right),
\label{eq:dirichlet_map}
\end{equation}
where \(\phi=(W,\boldsymbol b)\), \(W\in\mathbb{R}^{K\times K}\), and \(\boldsymbol b\in\mathbb{R}^K\).

The calibration parameters are obtained by minimizing a regularized negative log-likelihood objective:
\begin{equation}
\scriptsize
\hat\phi
=
\arg\min_{W,\boldsymbol b}
\left[
-
\sum_{n=1}^{N_{\mathrm{calib}}}
\log
\hat{\boldsymbol m}_{\phi}\!\left(\bar{\boldsymbol p}^{(n)}\right)[y^{(n)}]
+
\lambda_{\mathrm{id}}
\left(
\|W-I\|_F^2+\|\boldsymbol b\|_2^2
\right)
\right].
\label{eq:dirichlet_fit}
\end{equation}
We then approximate the conditional class posterior by
\begin{equation}
\mathbb{P}(y=k \mid \bar{\boldsymbol p})
\approx
\hat{\boldsymbol m}_{\hat\phi}(\bar{\boldsymbol p})[k].
\label{eq:posterior_estimate}
\end{equation}

Substituting the calibrated posterior estimate \eqref{eq:posterior_estimate} and the classwise reliability coefficients \(C_{\mathrm{VFL}}[k,k]\) into \eqref{eq:analytical_score} yields the test-time score
\begin{equation}
\hat s(\bar{\boldsymbol p})
=
\sum_{k=0}^{K-1}
\hat{\boldsymbol m}_{\hat\phi}(\bar{\boldsymbol p})[k]
\left(
C_{\mathrm{VFL}}[k,k]
-
\mathbf{1}\!\left[\hat y_{\mathrm{loc}}=k\right]
\right).
\label{eq:deployable_score}
\end{equation}
The final routing policy is therefore
\begin{equation}
\pi_{\tau}(\bar{\boldsymbol p})
=
\mathbf{1}\!\left[\hat s(\bar{\boldsymbol p})>\tau\right],
\label{eq:final_policy}
\end{equation}
that is, the sample is escalated whenever the estimated gain of switching to the collaborative Round--2 route exceeds the threshold \(\tau\).\footnote{Because the additional Round--2 communication cost is fixed per escalated sample in our protocol, this cost can be absorbed into the threshold \(\tau\). Thus, increasing \(\tau\) corresponds to requiring a larger expected accuracy gain before paying the communication cost. For a target escalation budget \(B\), \(\tau\) can be chosen on calibration data as the empirical \((1-B)\)-quantile of the routing scores, so that approximately a fraction \(B\) of samples is escalated.}

\begin{algorithm}[t]
\caption{Analytical Expected-Gain Routing}
\label{alg:analytical_routing}
\begin{algorithmic}[1]

\Require Calibration set $\mathcal{D}_{\mathrm{calib}}$, test set $\mathcal{D}_{\mathrm{test}}$, local classifiers $\{f_i\}_{i=1}^M$, encoders $\{E_i\}_{i=1}^M$, collaborative model $g_{\mathrm{VFL}}$, threshold $\tau$
\Ensure Routing policy $\pi_\tau(\bar{\boldsymbol p})$

\State \textbf{Offline calibration phase}
\State Estimate the Round--2 confusion matrix $C_{\mathrm{VFL}}$ on $\mathcal{D}_{\mathrm{calib}}$ via \eqref{eq:cvfl_estimator}
\State Fit Dirichlet calibration parameters $\hat{\phi}$ on $\mathcal{D}_{\mathrm{calib}}$ via \eqref{eq:dirichlet_fit}

\Statex
\State \textbf{Inference-time routing phase}
\For{each test sample $(\boldsymbol{x}_1,\ldots,\boldsymbol{x}_M)\in\mathcal{D}_{\mathrm{test}}$}
    \State Compute local posteriors $\{\boldsymbol p_i=f_i(\boldsymbol x_i)\}_{i=1}^M$
    \State Compute the pooled posterior via \eqref{eq:avg_pool}
    \State Compute the local prediction $\hat y_{\mathrm{loc}}$ via \eqref{eq:y_loc}
    \State Compute the calibrated posterior estimate $\hat{\boldsymbol m}_{\hat\phi}(\bar{\boldsymbol p})$ via \eqref{eq:posterior_estimate}
    \State Compute the analytical routing score $\hat s(\bar{\boldsymbol p})$ via \eqref{eq:deployable_score}
    \If{$\hat s(\bar{\boldsymbol p}) > \tau$}
        \State Compute embeddings $\{\boldsymbol e_i=E_i(\boldsymbol x_i)\}_{i=1}^M$
        \State Compute the collaborative posterior $\boldsymbol q=g_{\mathrm{VFL}}(\boldsymbol e_1,\ldots,\boldsymbol e_M)$ via \eqref{eq:vfl_posterior}
        \State Output $\hat y_{\mathrm{VFL}}=\argmax_k \boldsymbol q(k)$
    \Else
        \State Output $\hat y_{\mathrm{loc}}=\argmax_{k} \bar{\boldsymbol p}(k).$
    \EndIf
\EndFor

\end{algorithmic}
\end{algorithm}

\section{Evaluation Setting}
\label{sec:evaluation}

We evaluate the proposed routing rule on three classification benchmarks: CIFAR-10, CIFAR-100, and ModelNet40. Together, these datasets cover complementary settings, from standard 2D image classification to synthetic multi-view 3D object recognition. In each case, we construct a multi-client inference problem in which several clients observe partial views of the same underlying sample, while the server aggregates the communicated information and, when needed, applies the collaborative Round--2 predictor.

\subsection{Datasets}
\label{subsec:dataset}

\paragraph{CIFAR-10 and CIFAR-100}
For CIFAR-10 and CIFAR-100 \cite{cifar}, we create a four-client benchmark by partitioning each image into four non-overlapping quadrants taken from the same original image. Each quadrant is assigned to one client, so that the four clients receive complementary observations associated with the same class label. This provides a simple vertically partitioned setting in which no client has access to the full image.

\paragraph{ModelNet40}
For ModelNet40, we construct a two-client benchmark from the original 3D object models. Each object mesh is first normalized in scale and then rendered from two fixed viewpoints, producing a paired multi-view sample for the same underlying object. In the two-view setting used here, the rendered images correspond to a \emph{top} view and a \emph{diagonal} view, which are assigned to the two clients. The resulting pair provides complementary observations of the same 3D shape while preserving a shared class label.

\subsection{Baselines}
\label{subsec:baselines}

We compare against four baselines overall. The learned, confidence, and oracle baselines are shown in the main trade-off curves, while the L2D baseline is reported separately at matched escalation budgets.

\paragraph{Learned  ($\diffzeroone$) gain router}

We compare against a learned router that predicts the realized escalation gain. The router is a two-layer MLP trained on the calibration set to regress 
\[
G=\mathbf{1}[\hat y_{\mathrm{VFL}}=y]-\mathbf{1}[\hat y_{\mathrm{loc}}=y]
\]
using mean squared error. Its input consists of features $\boldsymbol z$ available after Round--1.  $\boldsymbol z$ includes the pooled posterior \(\bar{\boldsymbol p}\) together with uncertainty and disagreement summaries derived from the local and pooled predictions, such as the local entropies
\[
H(\boldsymbol p_i)=-\sum_{k=0}^{K-1} p_i(k)\log p_i(k), \qquad i=1,\dots,M,
\]
the entropy of the pooled posterior
\[
H(\bar{\boldsymbol p})=-\sum_{k=0}^{K-1} \bar p(k)\log \bar p(k),
\]
and the KL divergences between each local posterior and the pooled posterior
\[
D_{\mathrm{KL}}(\boldsymbol p_i \,\|\, \bar{\boldsymbol p})
=
\sum_{k=0}^{K-1} p_i(k)\log\frac{p_i(k)}{\bar p(k)},
\qquad i=1,\dots,M.
\]
These features provide a compact summary of Round--1 confidence and cross-client disagreement for training a gain predictor.

\paragraph{Confidence-based router}
This baseline escalates when the pooled Round--1 posterior is insufficiently confident, using the maximum pooled posterior mass as a scalar confidence score.

\paragraph{Learning to Defer-based router}
We also compare against a learning-to-defer (L2D) formulation in \cite{mozannar2020consistent}. Each client trains a \(K+1\)-class classifier, where the first \(K\) outputs correspond to the original task classes and the additional output corresponds to a defer/escalate action. The VFL model is treated as the downstream expert. After training the models of the Round--2 VFL route, the client models (i.e. the \(K+1\)-class classifiers)  are trained on the same training split using the VFL correctness indicator as a signal to tailor the loss. Specifically, for each training sample, the frozen Round--2 VFL route  predicts \(\hat y_{\mathrm{VFL}}\), and the indicator \(\mathbf{1}[\hat y_{\mathrm{VFL}}=y]\) determines whether deferral should be rewarded. When the Round--2 VFL route  is correct, the client is encouraged to assign probability both to the true class and to the defer output; when the Round--2 VFL route  is incorrect, the client is trained only to predict the true class locally. This corresponds to the loss
\[
\mathcal{L}_{\mathrm{L2D}}
=
-\alpha \log p_y^{+}
-
\mathbf{1}[\hat y_{\mathrm{VFL}}=y]\log p_{\mathrm{def}}^{+},
\]
where \(p_y^{+}\) is the client probability assigned to the true class, \(p_{\mathrm{def}}^{+}\) is the probability assigned to the defer output, and \(\alpha\) controls the classification term. At inference time, the server averages the \(K+1\)-dimensional posteriors. Let
\[
\bar{\boldsymbol p}^{+}\in\mathbb{R}^{K+1}
\]
denote this averaged posterior, and let \(\bar p_{\mathrm{def}}^{+}\) be its defer component. The routing score is
\[
s_{\mathrm{L2D}}
=
\bar p_{\mathrm{def}}^{+}
-
\max_{k=0,\dots,K-1}\bar p^{+}(k),
\]
and escalation occurs when \(s_{\mathrm{L2D}}>\tau\). The canonical decision here corresponds to taking the argmax over the \(K+1\) outputs and escalating whenever the defer class is selected. We compare our analytical model with this baseline separately at matched escalation budgets for different values of $\alpha$ in Table \ref{tab:horizontal-budget-routing}.

\paragraph{Oracle ($\diffzeroone$) router.}
This reference policy has access to the realized gain \(G\) and therefore represents an upper bound on routing performance under a given escalation budget. It is not deployable in practice.

Table~\ref{tab:routing_rules} summarizes the scalar routing score used by each method and the corresponding escalation rule.

\begin{table*}[t]
\centering
\caption{Routing scores and escalation rules used in evaluation.}
\label{tab:routing_rules}

\begin{tabular*}{\textwidth}{@{\extracolsep{\fill}}lll@{}}
\hline
\textbf{Method} & \textbf{Routing score} & \textbf{Escalate if} \\
\hline

Proposed (Analytical) &
$\hat s(\bar{\boldsymbol p})$ &
$\hat s(\bar{\boldsymbol p})>\tau$ \\

\diffzeroone &
$g_{\psi}(\boldsymbol z)
\approx
\mathbf{1}[\hat y_{\mathrm{VFL}}=y]
-
\mathbf{1}[\hat y_{\mathrm{loc}}=y]$ &
$g_{\psi}(\boldsymbol z)>\tau$ \\

Confidence &
$-\max_k \bar{\boldsymbol p}(k)$ &
$\max_k \bar{\boldsymbol p}(k) < \gamma,\ \gamma\in[0,1]$ \\

\cite{mozannar2020consistent} &
$\bar p_{\mathrm{def}}^{+}
-
\max_{k=0,\dots,K-1}\bar p^{+}(k)$ &
$\bar p_{\mathrm{def}}^{+}
-
\max_{k=0,\dots,K-1}\bar p^{+}(k)>\tau$ \\

\diffzeroone\ Oracle &
$G=
\mathbf{1}[\hat y_{\mathrm{VFL}}=y]
-
\mathbf{1}[\hat y_{\mathrm{loc}}=y]$ &
beneficial $\rightarrow$ neutral $\rightarrow$ harmful \\

\hline
\end{tabular*}
\end{table*}
\subsection{Training and Calibration Protocol}
\label{subsec:protocol}

For each dataset, the available training data is split into two disjoint subsets. The first is used to train both the client-side local predictors and the Round--2 collaborative VFL model. The second is held out for calibration and router construction. This calibration subset is used to fit the Dirichlet calibration map for the pooled Round--1 posterior and to estimate the classwise Round--2 reliability coefficients used by the analytical score. The same subset is also used to fit the learned $\diffzeroone$ gain router, so that the analytical router and learned routing baseline rely on the same data split.

The Round--1 predictor is built from local classifiers that operate only on the client's own partial observation. The Round--2 predictor uses client-specific encoders together with a server-side fusion head. In our implementation, the Round--2 encoders are based on ResNet-18 backbones \cite{resnet18}, followed by a prediction head at the server. Since both the local predictors and the collaborative model are trained on the same training subset, differences between the two routes reflect their modeling and communication capabilities rather than unequal access to training data.

\subsection{Inference-Time View Degradation}
\label{subsec:varying_local_accuracy}

To examine how routing behavior changes as the quality of client views degrades, we introduce controlled input corruption at inference time. Specifically, we add zero-mean Gaussian noise with fixed standard deviation $\sigma=0.25$ to the normalized inputs of each client for a selected subset of classes during test-time evaluation. No corruption is applied during training or calibration: the local predictors, the Round--2 collaborative model, and the posterior calibration map are all learned from clean data. This setup therefore introduces a controlled mismatch between model construction and deployment conditions.

The noise is applied consistently across the inference pipeline, affecting both the Round--1 local predictors and the client encoders used by the Round--2 VFL model. As the number of affected classes increases, a larger fraction of the input space is subject to degraded local observations, making the routing problem progressively more challenging.

This is intended to emulate test-time heterogeneity, where the informativeness of clients views may degrade unevenly across the data--for example due to occlusion, blur, illumination variation, truncation, or sensor noise. The goal is to assess whether the proposed routing rule remains effective under increasingly adverse deployment conditions.
\subsection{Model Architectures}
\label{subsec:model_architectures}

\paragraph{Round--1 Local Models}
In all experiments,  clients use the same lightweight local classifier\footnote{Exception applies to the \cite{mozannar2020consistent} baseline experiment, as clients' local models differ in their output layer which outputs $K$ label posteriors in addition to a defer posterior, as described in \autoref{subsec:baselines}.} in Round--1: a small convolutional network with four strided $3\times 3$ convolutional blocks, one additional same-resolution refinement block, global average pooling, dropout, and a final linear layer. This design keeps the first route intentionally simple.

\paragraph{Round--2 VFL Local and Server Models}
Round--2 uses one ResNet-18 encoder at each client to produce the embeddings, followed by a fusion head at the server. The fusion head projects client embeddings into a common latent space, computes sample-dependent client gates, and classifies from a representation that combines the gated fusion with explicit interaction features.

\paragraph{Posterior Calibration}
Across all datasets, the pooled Round--1 posterior is post-hoc calibrated with a multiclass Dirichlet calibrator using a held-out calibration dataset\footnote{For fairness, the same calibration split is used to train the learned $\diffzeroone$ router.}.

Dataset-specific training, calibration, and routing hyperparameters are summarized in the Appendix.

\subsection{Evaluation Metrics}
\label{subsec:metrics}

Let
\[
\pi_{\tau}(\bar{\boldsymbol p})
=
\mathbf{1}\!\left[\hat s(\bar{\boldsymbol p})>\tau\right]
\]
denote the analytical routing policy under threshold $\tau$, where
$\hat s(\bar{\boldsymbol p})$ is the deployable score defined in
\eqref{eq:deployable_score}. Here,
\(\pi_{\tau}(\bar{\boldsymbol p})=1\) means that the sample is escalated
to the VFL route, while \(\pi_{\tau}(\bar{\boldsymbol p})=0\) means that
the Round--1 local prediction is retained. The final prediction is
\[
\hat y^{\pi_\tau}
=
\begin{cases}
\hat y_{\mathrm{VFL}}, & \pi_\tau(\bar{\boldsymbol p})=1,\\
\hat y_{\mathrm{loc}}, & \pi_\tau(\bar{\boldsymbol p})=0.
\end{cases}
\]
Varying $\tau$ traces different communication--accuracy operating points.
We report the following metrics:
\paragraph{System accuracy.}
\[
\mathrm{Acc.}(\tau)
\triangleq
\mathbb{E}\!\left[
\mathbf{1}\!\left[\hat y^{\pi_\tau}=y\right]
\right].
\]

\paragraph{Escalation rate.}
\[
\mathrm{ER}(\tau)
\triangleq
\mathbb{E}\!\left[
\pi_\tau(\bar{\boldsymbol p})
\right].
\]

Sweeping \(\tau\) on the test set yields the trade-off curve
\[
\bigl(\mathrm{ER}(\tau),\mathrm{Acc.}(\tau)\bigr).
\]
For reference, we also report the two endpoint policies: never escalating,
which corresponds to local-only prediction, and always escalating, which
corresponds to Round--2 VFL route  prediction.

To assess whether escalations are well targeted, we report
\begin{align}
\mathrm{EP}(\tau)
&\triangleq
\mathbb{E}\!\left[
\mathbf{1}[G=+1]
\mid
\pi_\tau(\bar{\boldsymbol p})=1
\right], \nonumber\\
\mathrm{NE}(\tau)
&\triangleq
\mathbb{E}\!\left[
\mathbf{1}[G=0]
\mid
\pi_\tau(\bar{\boldsymbol p})=1
\right]. \nonumber
\end{align}

\(\mathrm{EP}\) is the escalation precision: among escalated samples,
it measures the fraction for which VFL corrects a Round--1 error.
\(\mathrm{NE}\) is the neutral escalation share: among escalated samples,
it measures the fraction for which escalation does not change correctness.
Thus, high \(\mathrm{EP}\) indicates that escalations are often beneficial,
whereas high \(\mathrm{NE}\) indicates that many VFL calls are unnecessary.

\section{Results}

\begin{figure*}[t]
\centering

\begin{minipage}[t]{0.325\textwidth}
\centering

\includegraphics[width=\linewidth]{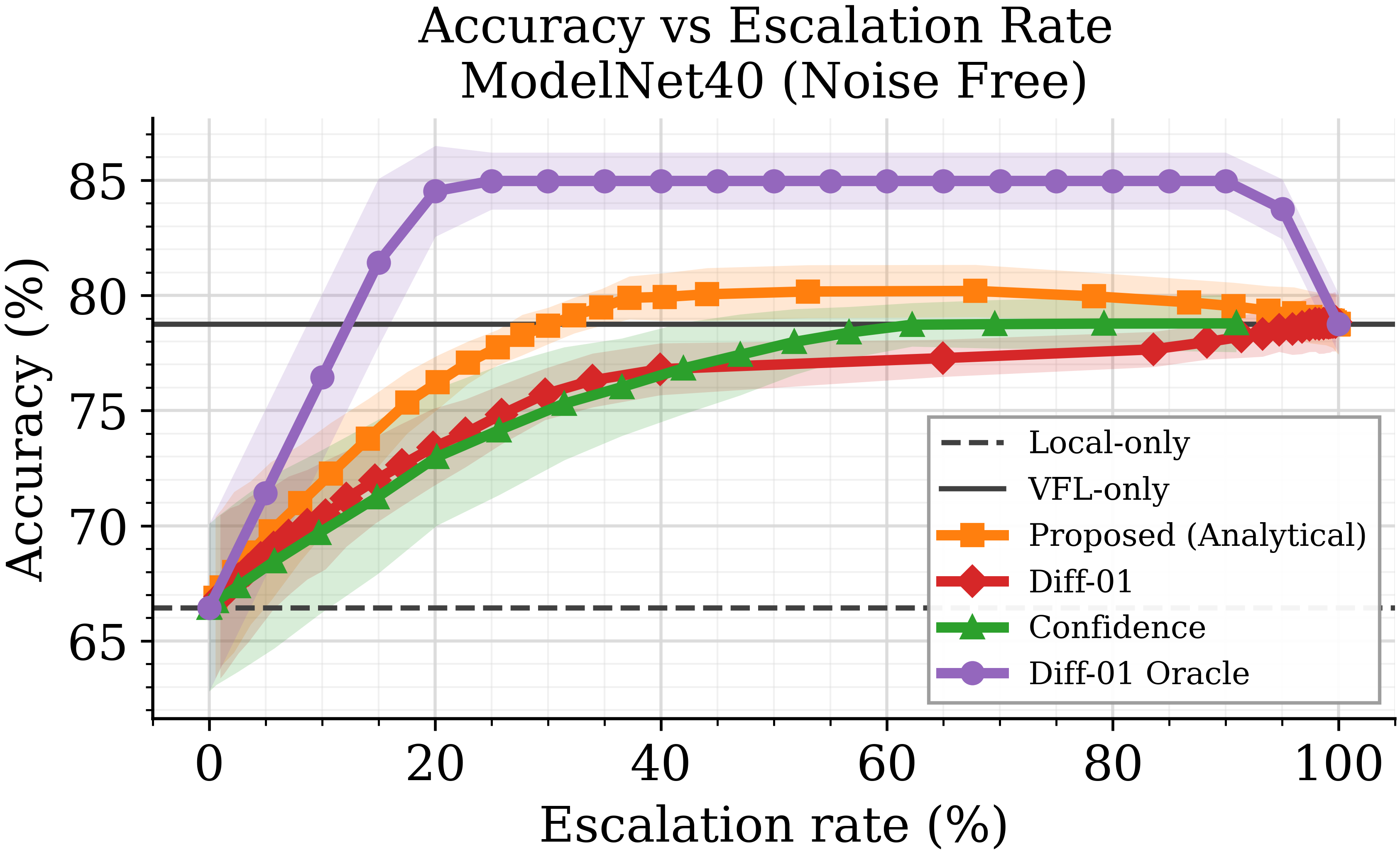}

\vspace{0.3em}

\includegraphics[width=\linewidth]{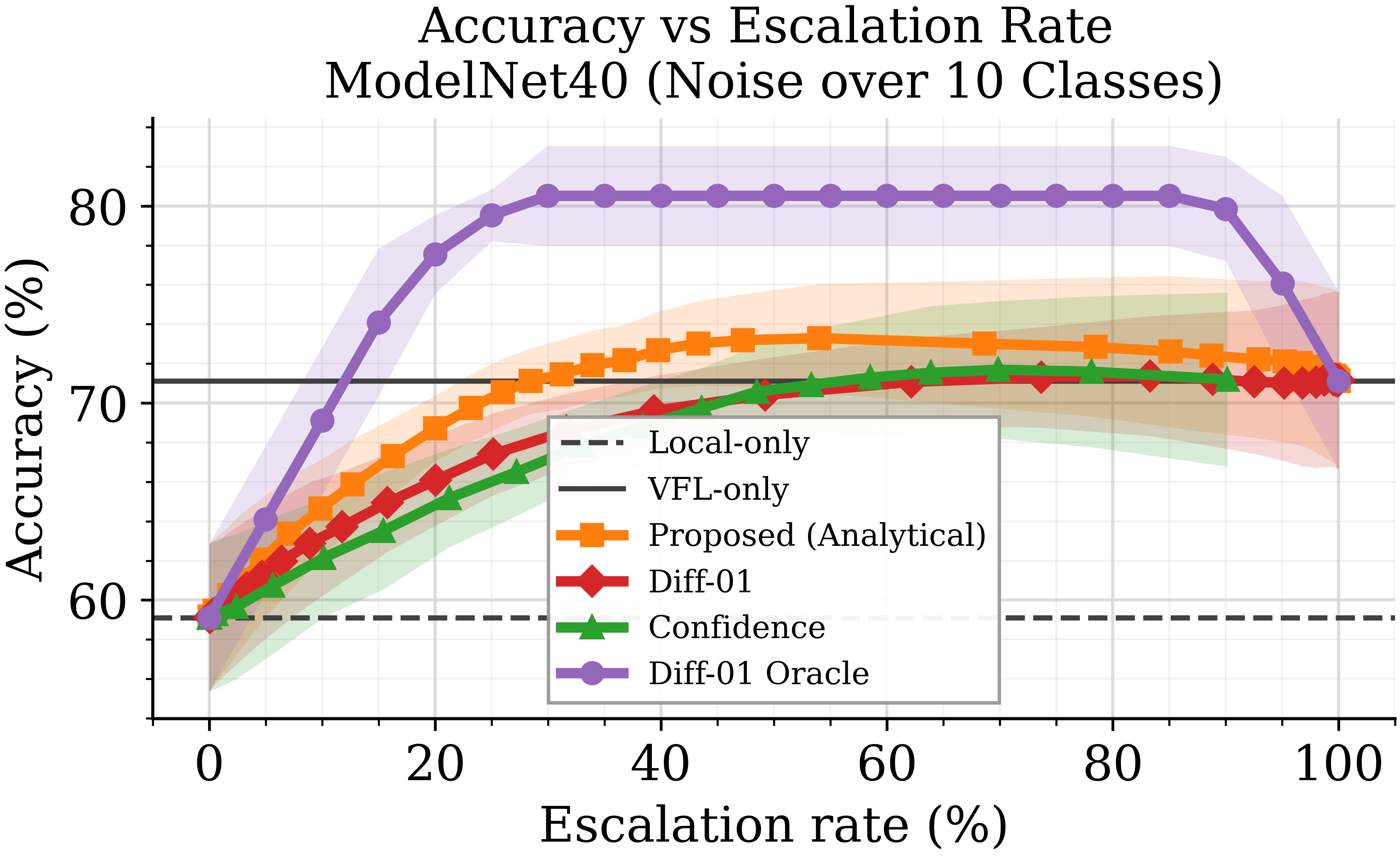}
\vspace{0.3em}

\includegraphics[width=\linewidth]{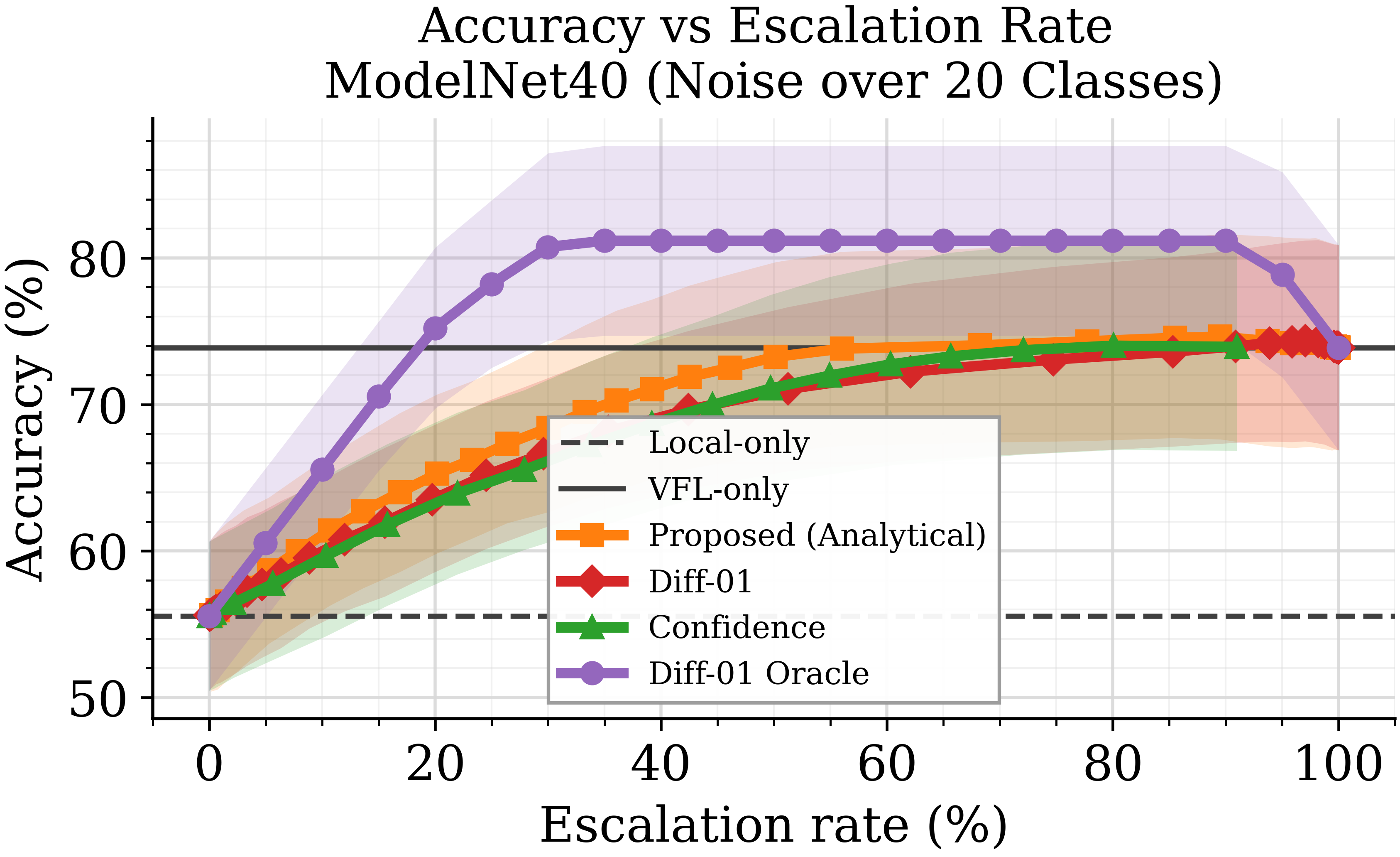}

\end{minipage}
\hfill
\begin{minipage}[t]{0.325\textwidth}
\centering

\includegraphics[width=\linewidth]{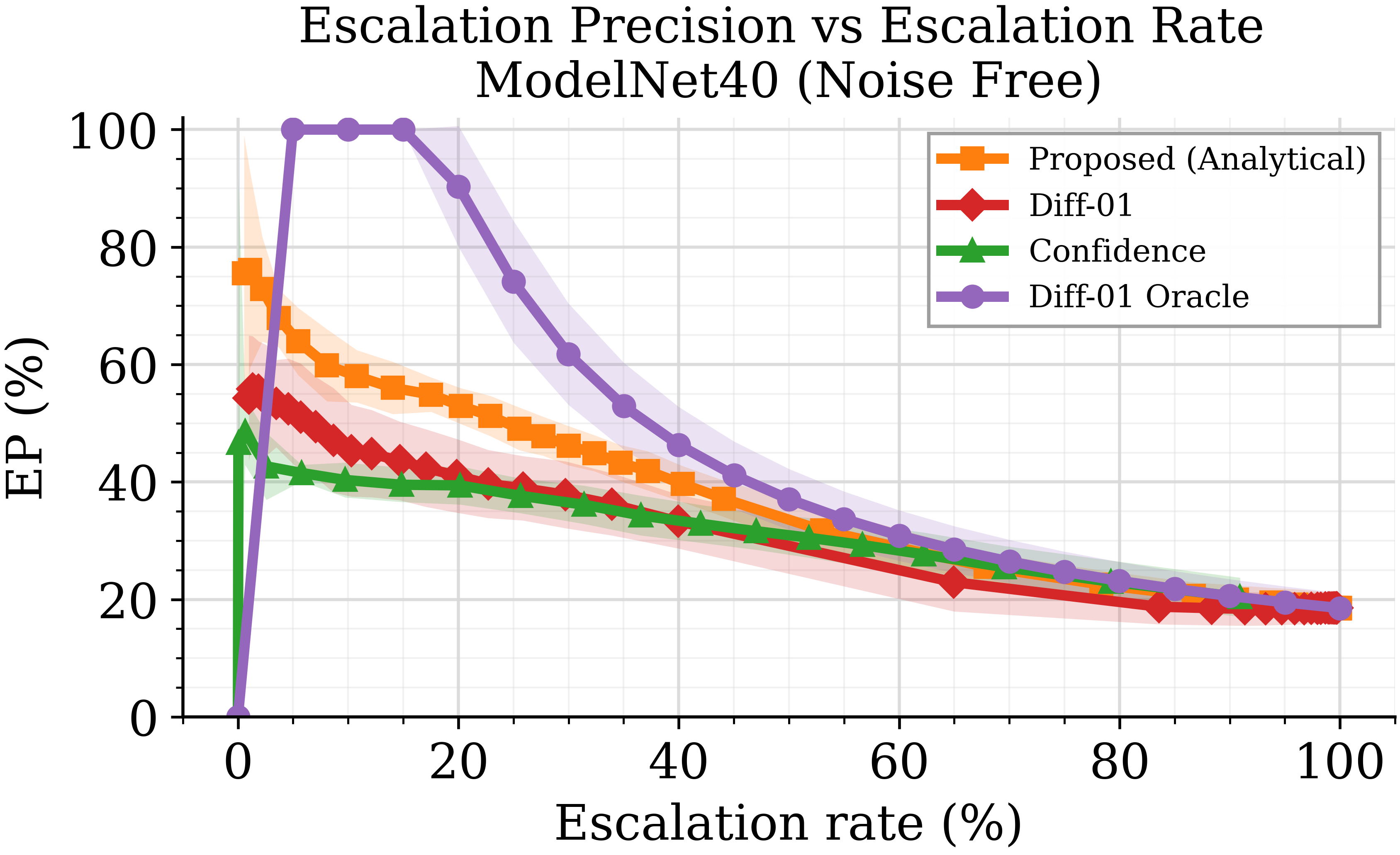}

\vspace{0.3em}

\includegraphics[width=\linewidth]{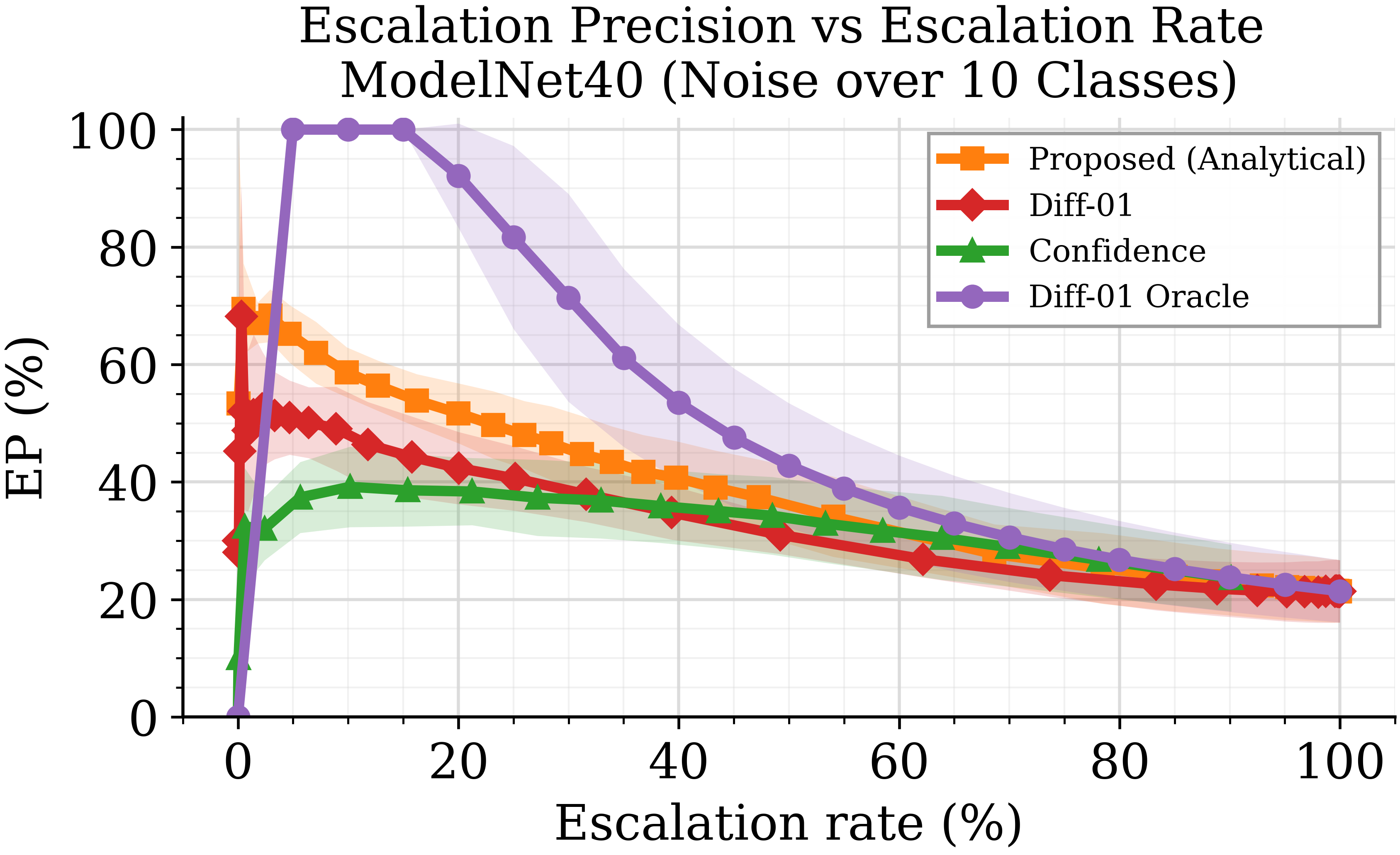}
\vspace{0.3em}

\includegraphics[width=\linewidth]{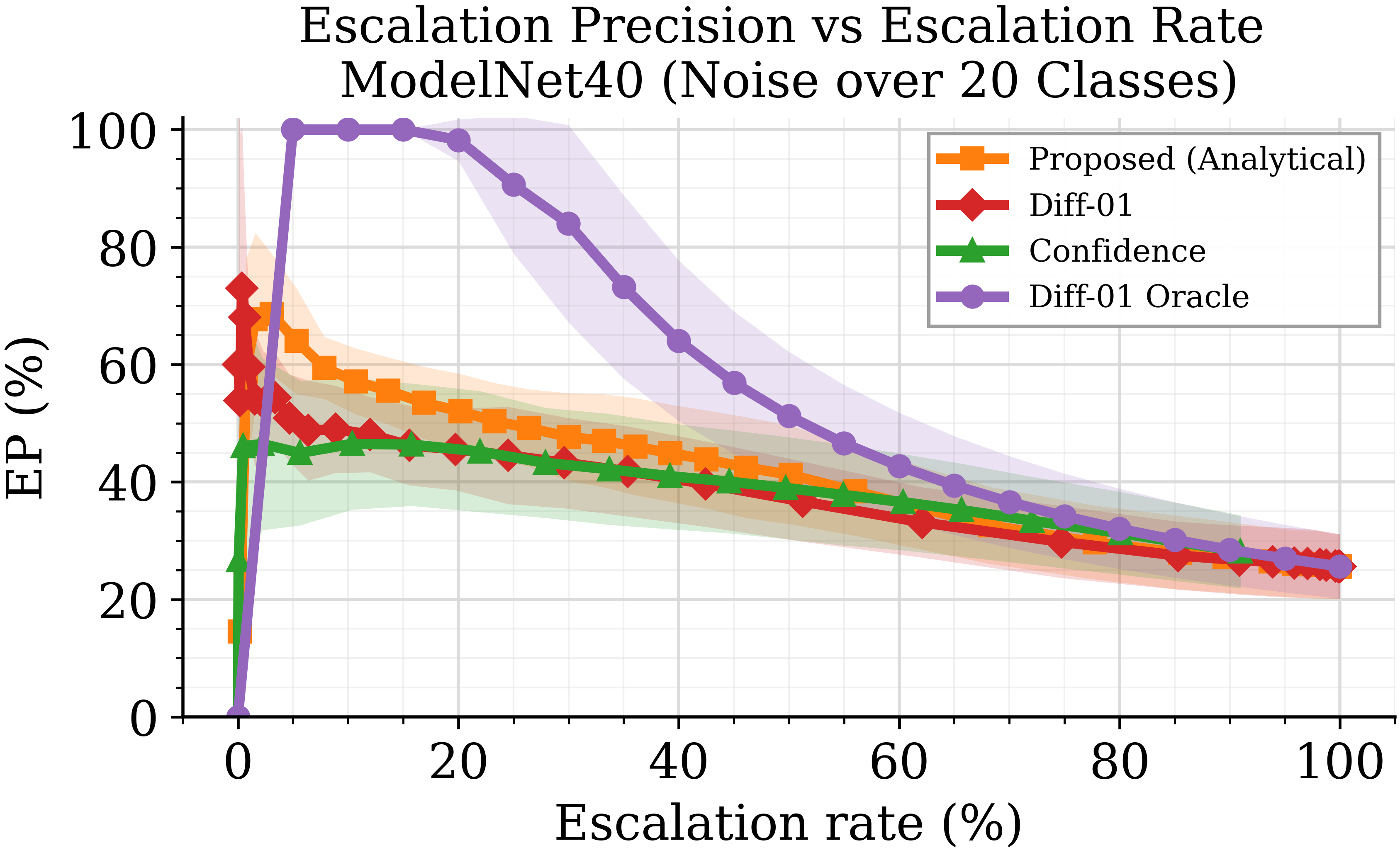}

\end{minipage}
\hfill
\begin{minipage}[t]{0.325\textwidth}
\centering

\includegraphics[width=\linewidth]{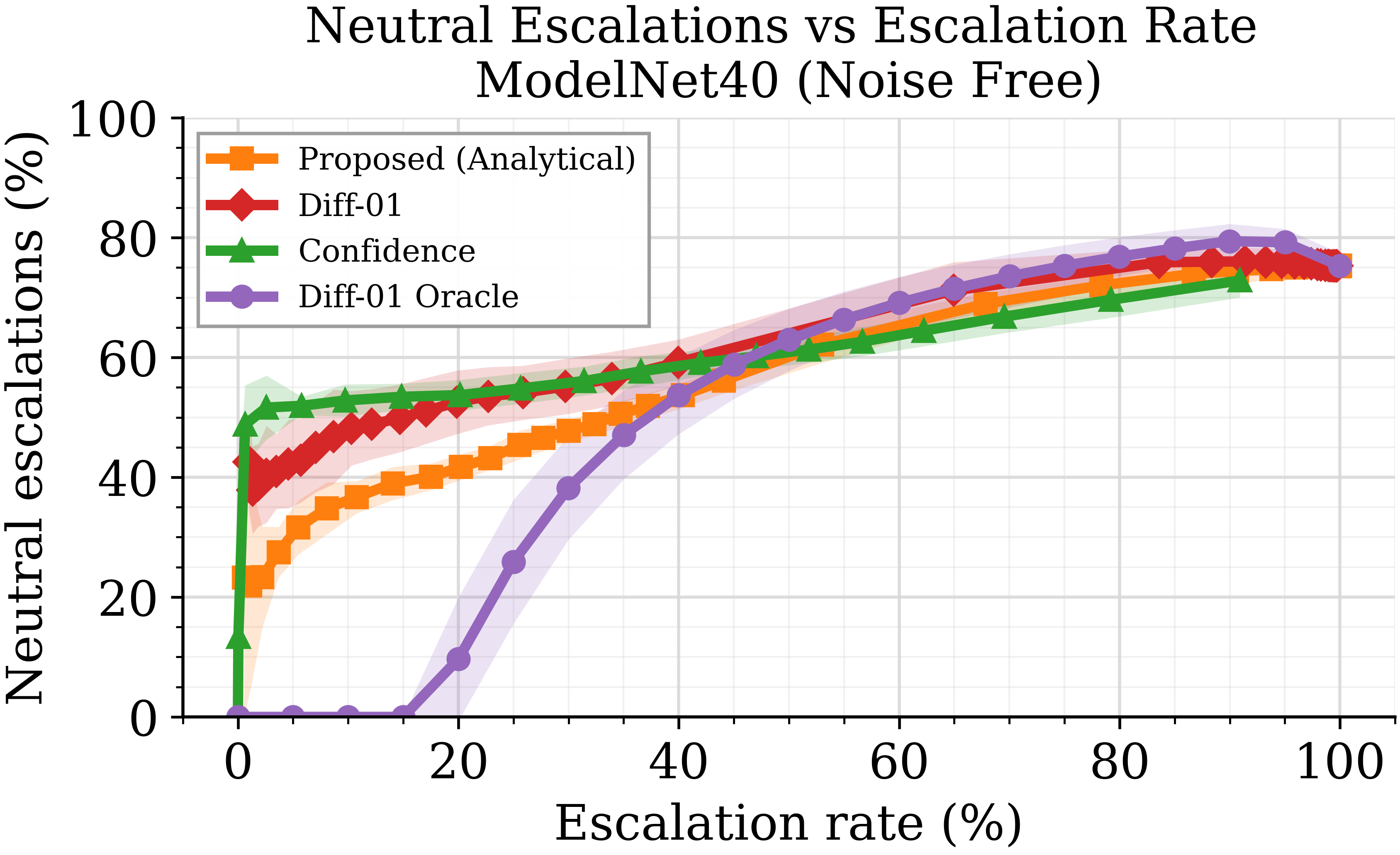}

\vspace{0.3em}

\includegraphics[width=\linewidth]{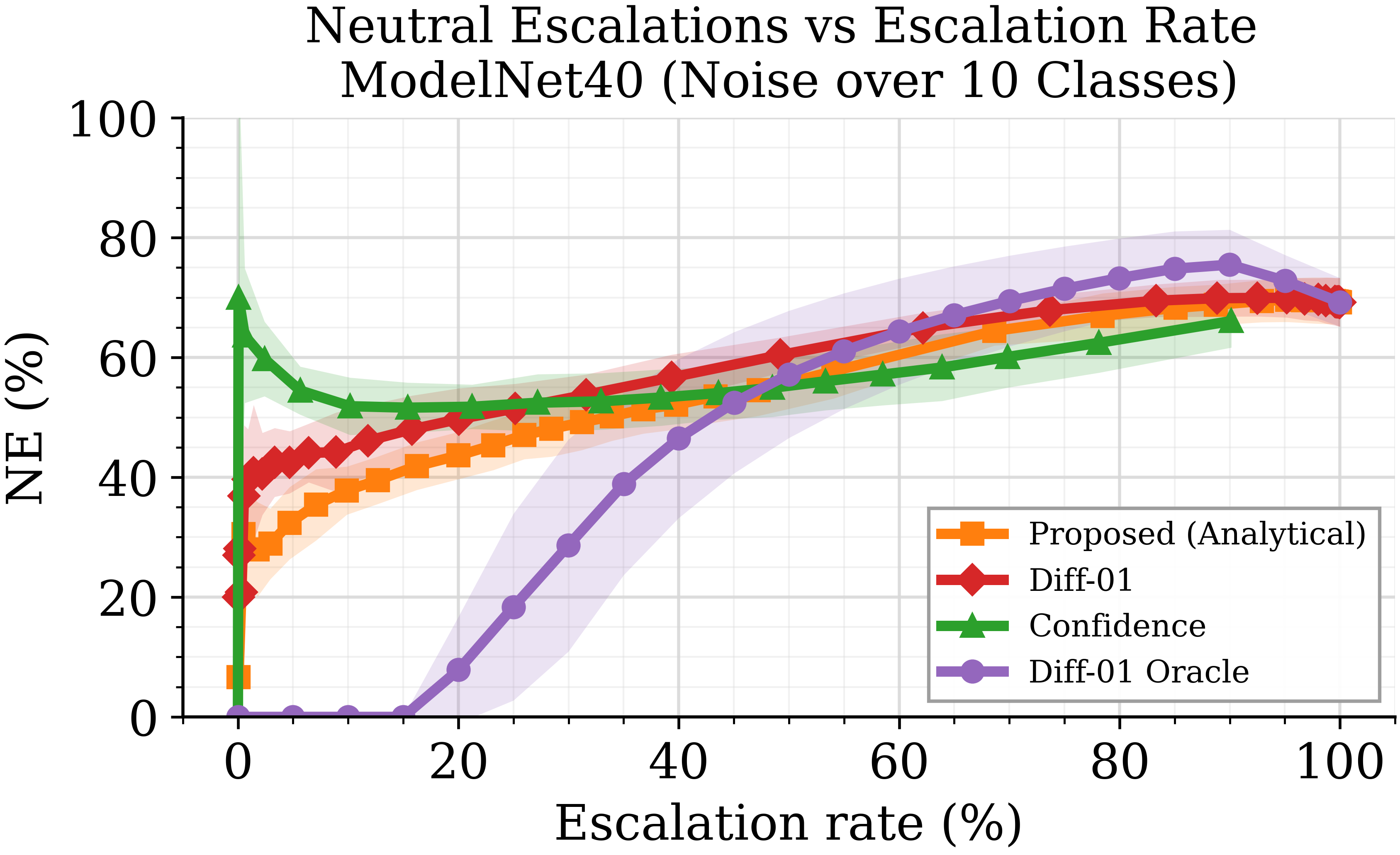}
\vspace{0.3em}

\includegraphics[width=\linewidth]{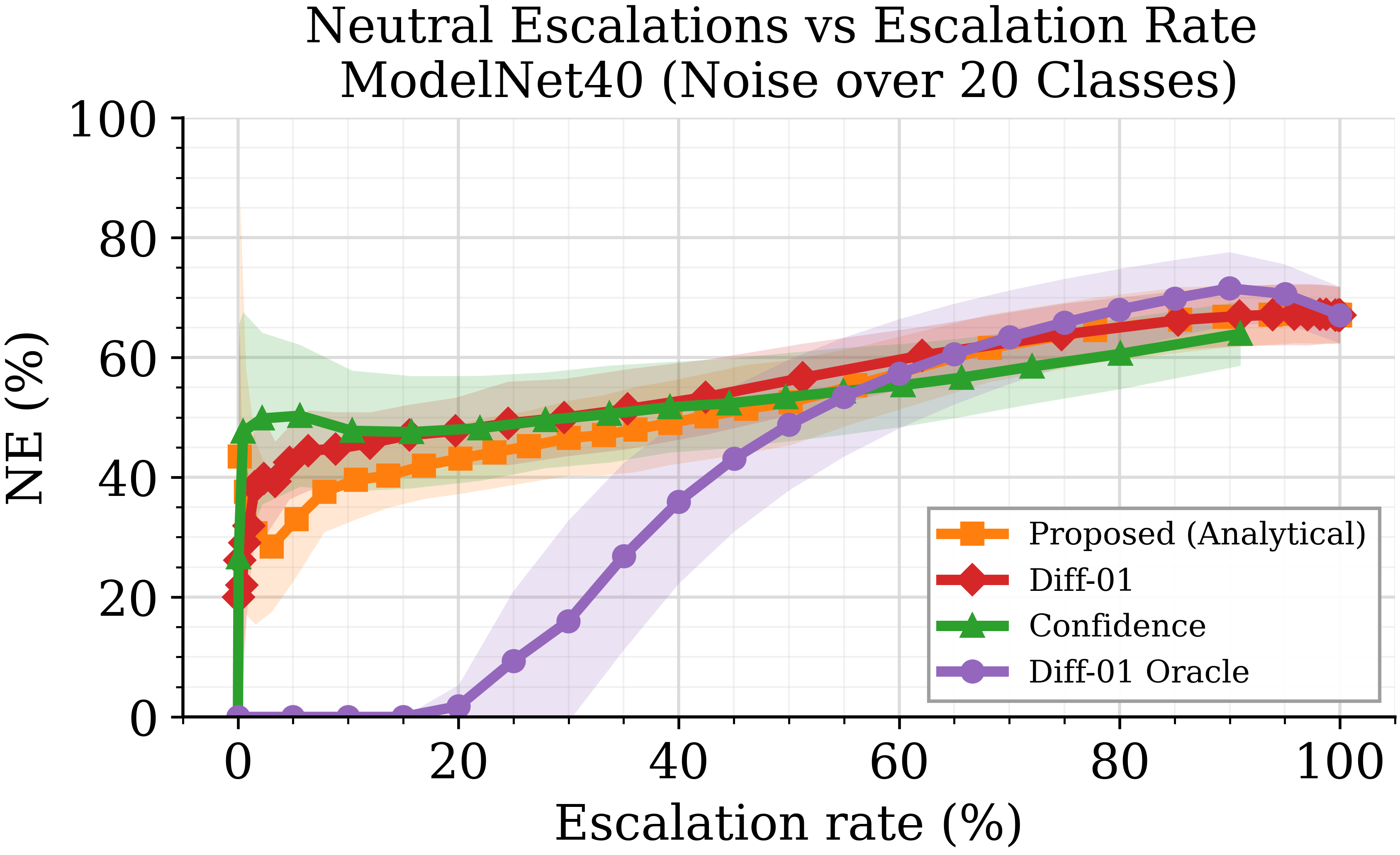}

\end{minipage}
\caption{Comparison of the proposed analytical expected-gain router, the learned router, the oracle, and a confidence-based baseline across three inference-time noise regimes on the ModelNet40 dataset. Each row corresponds to a different inference-time noise regime. Columns report accuracy, escalation precision (EP), and the proportion of neutral escalations (NE), all shown as functions of the escalation rate.}
\label{fig:resvsbase}
\end{figure*}

We first analyze ModelNet40 in detail, then summarize the CIFAR results. \autoref{fig:resvsbase} summarizes average (over 5 runs) test performance on ModelNet40 across the three inference-time Gaussian-noise regimes. Among the practical routing rules, the analytical expected-gain router performs best overall. Across operating points, it reaches a higher peak accuracy than both endpoint policies---local-only and always-escalate VFL---in all three noise regimes. By contrast, the $\diffzeroone$ learned router is consistently weaker. This suggests that the calibrated expected-gain estimate already captures most of the useful routing signal, allowing the router to prioritize samples that are likely to benefit from escalation while avoiding those that are neutral or harmful. This  pattern is reflected in the EP and NE curves.

In particular, in the noise-free setting, the pooled Round--1 local predictor reaches roughly $67\%$ accuracy; this drops to about $59\%$ and $56\%$ as noise is applied to 10 and 20 classes, respectively. The always-escalate Round--2 VFL route follows a similar trend, decreasing from about $79\%$ in the noise-free setting to around  $71$--$74\%$ under the two noisy regimes, while remaining consistently stronger than Round--1. This confirms that escalation remains beneficial in principle, even under substantial degradation of the clients views.

A second consistent pattern is that most test samples are neutral with respect to escalation: the Oracle curve saturates early (around $20$--$25\%$ on average), indicating that only a subset of examples truly benefit from invoking Round--2. As a consequence, very low routing thresholds lead to over-escalation, since many Round--2 calls do not change the final prediction and may occasionally even hurt it.

The most effective operating points for our method occur at moderate escalation rates, approximately \(40\)--\(60\%\) in the noise-free setting and when noise is applied to 10 classes. This corresponds to escalating roughly half of the test samples rather than operating near either endpoint. This behavior is consistent with the intended role of routing: the router filters out many neutral escalations, preserves most beneficial escalations, and avoids harmful escalations until high escalation rates force their inclusion. As the escalation rate increases further, the accuracy of our method gradually declines and eventually converges to that of the always-escalate VFL baseline, both in the noise-free case and under noise affecting $10$ classes. This decline reflects the point at which the router is forced to escalate samples that were previously held out precisely because they are harmful.

Importantly, the same qualitative behavior is observed for the Oracle strategy: once all beneficial samples have already been escalated and harmful samples can no longer be avoided, additional escalation inevitably includes detrimental cases, leading to a degradation in overall accuracy.

The confidence-based baseline improves over local-only prediction, but remains consistently less effective than the proposed analytical router, indicating that confidence alone is not a sufficiently informative signal for escalation.
The Oracle performance remains better than all deployable methods. The remaining gap between the oracle and the analytical router indicates clear room for improving sample selection.

\begin{table*}[t]
\centering
\caption{
Comparison of the proposed analytical router with \cite{mozannar2020consistent} for $\alpha=0.5,1,2,3$.
Bold indicates best performance per budget (higher is better for Acc./EP, lower for NE).
}
\label{tab:horizontal-budget-routing}
\small
\renewcommand{\arraystretch}{1.1}

\begin{tabular*}{\textwidth}{@{\extracolsep{\fill}} 
p{2.8cm} p{2.2cm} c c c c c c @{}}

\toprule
\textbf{Method} & \textbf{Metric (mean $\pm$ std)} &
\shortstack{0\%\\Local} & 20\% & 40\% & 60\% & 80\% &
\shortstack{100\%\\VFL} \\
\midrule

Proposed (Analytical) & Acc.
& 65.22 $\pm$ 4.53
& \textbf{75.40 $\pm$ 5.09}
& \textbf{78.29 $\pm$ 4.40}
& \textbf{78.30 $\pm$ 4.61}
& 77.60 $\pm$ 4.80
& 74.67 $\pm$ 9.21 \\

\cite{mozannar2020consistent} ($\alpha=0.5$)
& Acc.
& 63.98 $\pm$ 4.22
& 69.76 $\pm$ 3.64
& 73.53 $\pm$ 3.19
& 76.29 $\pm$ 2.85
& 77.18 $\pm$ 2.51
& 74.67 $\pm$ 9.21 \\

\cite{mozannar2020consistent} ($\alpha=1$)
& Acc.
& 64.57 $\pm$ 4.89
& 69.71 $\pm$ 5.49
& 73.24 $\pm$ 5.02
& 75.49 $\pm$ 4.71
& 76.49 $\pm$ 4.74
& 74.67 $\pm$ 9.21 \\

\cite{mozannar2020consistent} ($\alpha=2$)
& Acc.
& \textbf{65.55 $\pm$ 3.48}
& 70.92 $\pm$ 3.78
& 74.81 $\pm$ 3.29
& 77.02 $\pm$ 2.19
& \textbf{77.84 $\pm$ 1.45}
& 74.67 $\pm$ 9.21 \\

\cite{mozannar2020consistent} ($\alpha=3$)
& Acc.
& 63.11 $\pm$ 0.85
& 68.82 $\pm$ 1.94
& 73.48 $\pm$ 1.79
& 76.27 $\pm$ 1.96
& 77.22 $\pm$ 2.34
& 74.67 $\pm$ 9.21 \\

\midrule

Proposed (Analytical) & EP
& 0.00 $\pm$ 0.00
& \textbf{54.43 $\pm$ 5.15}
& \textbf{39.35 $\pm$ 5.01}
& 28.01 $\pm$ 3.12
& 21.44 $\pm$ 2.86
& 17.77 $\pm$ 2.88 \\

\cite{mozannar2020consistent} ($\alpha=0.5$)
& EP
& 0.00 $\pm$ 0.00
& 35.06 $\pm$ 5.71
& 30.78 $\pm$ 5.57
& 27.72 $\pm$ 5.43
& 23.99 $\pm$ 5.03
& 20.18 $\pm$ 3.42 \\

\cite{mozannar2020consistent} ($\alpha=1$)
& EP
& 0.00 $\pm$ 0.00
& 34.67 $\pm$ 6.33
& 30.28 $\pm$ 5.58
& 26.47 $\pm$ 5.41
& 22.98 $\pm$ 5.33
& 19.46 $\pm$ 4.27 \\

\cite{mozannar2020consistent} ($\alpha=2$)
& EP
& 0.00 $\pm$ 0.00
& 34.35 $\pm$ 2.66
& 30.69 $\pm$ 2.23
& 26.89 $\pm$ 2.59
& 23.18 $\pm$ 2.96
& 19.29 $\pm$ 2.51 \\

\cite{mozannar2020consistent} ($\alpha=3$)
& EP
& 0.00 $\pm$ 0.00
& 36.54 $\pm$ 3.03
& 33.69 $\pm$ 2.97
& \textbf{29.83 $\pm$ 2.81}
& \textbf{25.33 $\pm$ 2.30}
& 20.76 $\pm$ 1.25 \\

\midrule

Proposed (Analytical) & NE
& 0.00 $\pm$ 0.00
& \textbf{41.19 $\pm$ 3.52}
& \textbf{53.65 $\pm$ 4.02}
& 65.87 $\pm$ 2.23
& 72.56 $\pm$ 2.37
& 73.91 $\pm$ 4.05 \\

\cite{mozannar2020consistent} ($\alpha=0.5$)
& NE
& 0.00 $\pm$ 0.00
& 59.14 $\pm$ 5.35
& 62.49 $\pm$ 5.30
& 65.34 $\pm$ 5.27
& 68.71 $\pm$ 5.30
& 72.30 $\pm$ 4.27 \\

\cite{mozannar2020consistent} ($\alpha=1$)
& NE
& 0.00 $\pm$ 0.00
& 57.86 $\pm$ 5.12
& 62.14 $\pm$ 4.96
& 65.71 $\pm$ 5.18
& 69.04 $\pm$ 5.32
& 71.18 $\pm$ 6.02 \\

\cite{mozannar2020consistent} ($\alpha=2$)
& NE
& 0.00 $\pm$ 0.00
& 58.49 $\pm$ 1.78
& 61.97 $\pm$ 1.50
& 65.45 $\pm$ 2.20
& 69.05 $\pm$ 2.50
& 73.55 $\pm$ 1.96 \\

\cite{mozannar2020consistent} ($\alpha=3$)
& NE
& 0.00 $\pm$ 0.00
& 56.39 $\pm$ 2.25
& 59.22 $\pm$ 2.27
& \textbf{62.66 $\pm$ 2.10}
& \textbf{67.15 $\pm$ 2.05}
& 72.24 $\pm$ 0.82 \\

\bottomrule
\end{tabular*}
\end{table*}
We further compare against the L2D baseline of \cite{mozannar2020consistent}. We report the comparison separately at matched escalation budgets in \autoref{tab:horizontal-budget-routing} while accounting for the baseline performance for different values of $\alpha$. The 0\% escalation (local-only) point shows similar performance among Round--1 predictions. As expected, EP and NE are both \(0\%\) at this point because no samples are escalated.

The difference in performance becomes clearer once escalation is allowed. At \(20\%\) escalation, the proposed router reaches \(75.40\%\) accuracy, compared with \(70.92\%\) for $\alpha=2$ in the L2D  baseline. Equivalently, our proposed router captures a much larger fraction of beneficial escalations, with an absolute EP improvement of about 18 percentage points over the strongest L2D variant at that budget. At \(40\%\) escalation, the same pattern persists. At higher escalation rates, the gap naturally narrows as more samples are sent  through the VFL-based route regardless of the router score (i.e. threshold $\tau$ becomes small). Still, the proposed router keeps the better end-to-end accuracy at \(60\%\). Finally, at \(80\%\),  some L2D variants match or slightly exceed the analytical router on these metrics. However, this occurs only at a much larger communication budget, whereas the analytical router already reaches comparable or better accuracy at \(40\%\)--\(60\%\) escalation.

At \(100\%\) escalation, both methods reduce to pure VFL inference, and therefore obtain the same final accuracy, \(74.67\%\). The EP and NE values, however, do not have to match at this endpoint. They are computed from the realized gain with respect to each method's own Round--1 local prediction models performance. Unlike other experiments which differ only in the routing strategy and share the same Round--1 and Round--2 models; here, the analytical router uses the pooled \(K\)-class local predictors as its local baseline, whereas the L2D baseline uses the prediction of its trained \(K+1\)-class client model. Thus, even though all final predictions come from the same VFL-based route at \(100\%\) escalation, the underlying per-sample realized gain can differ, which explains why EP and NE are different.

\begin{figure*}[t]
\centering

\begin{minipage}[t]{0.325\textwidth}
\centering

\includegraphics[width=\linewidth]{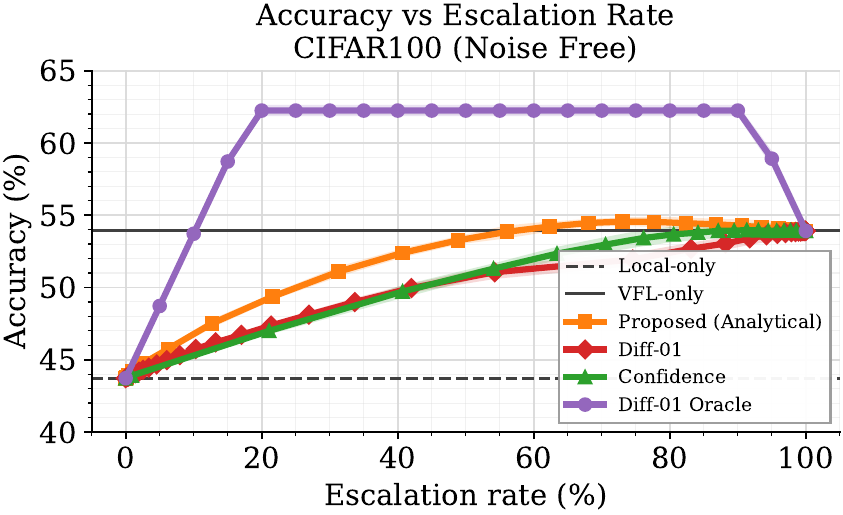}

\vspace{0.3em}

\includegraphics[width=\linewidth]{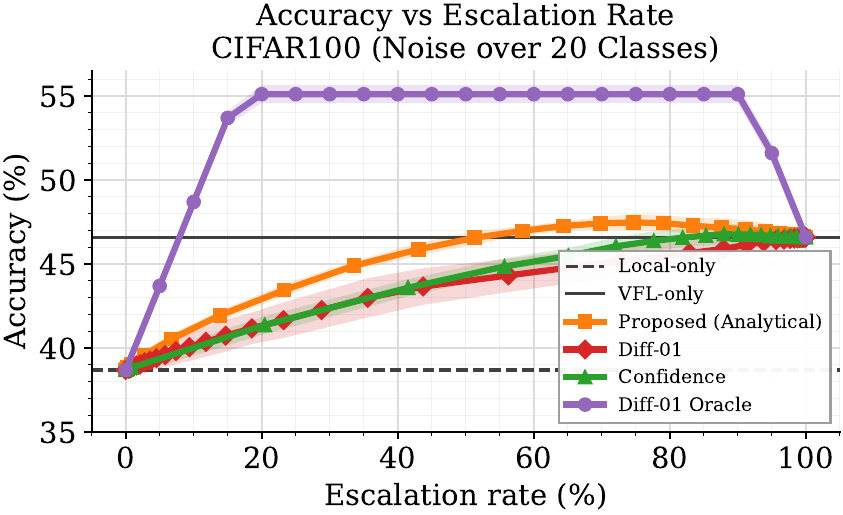}
\end{minipage}
\hfill
\begin{minipage}[t]{0.325\textwidth}
\centering

\includegraphics[width=\linewidth]{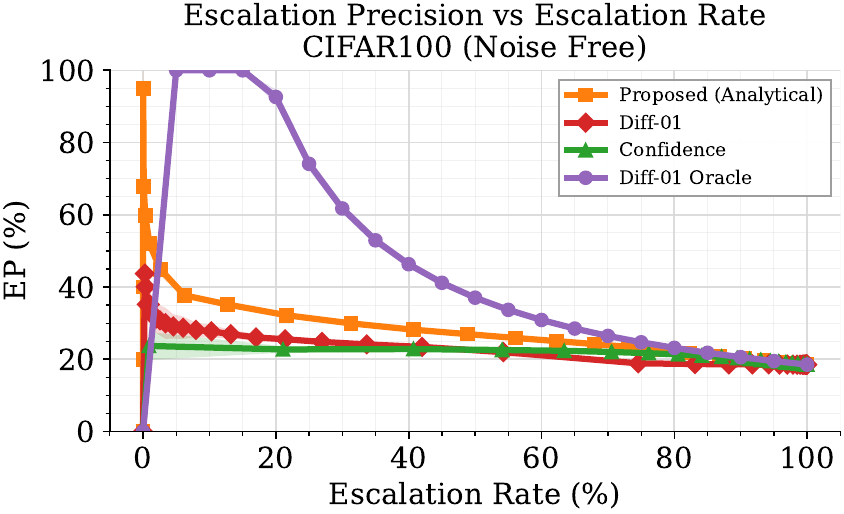}

\vspace{0.3em}

\includegraphics[width=\linewidth]{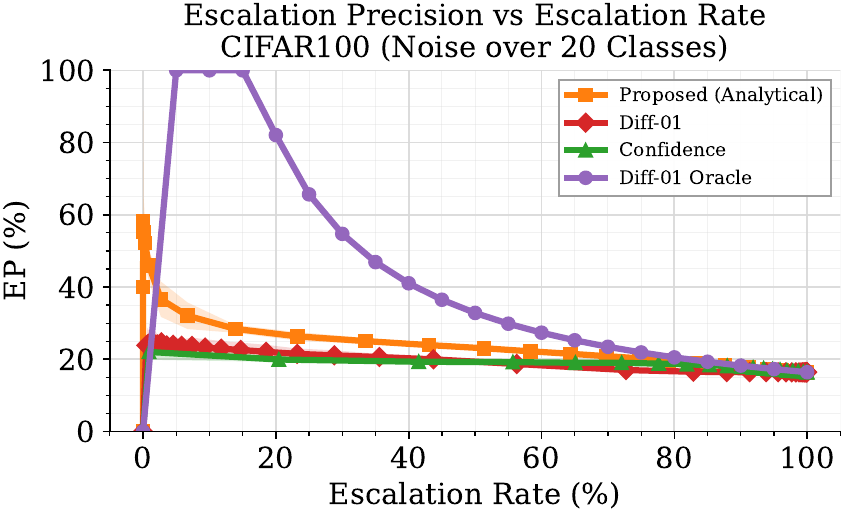}
\end{minipage}
\hfill
\begin{minipage}[t]{0.325\textwidth}
\centering

\includegraphics[width=\linewidth]{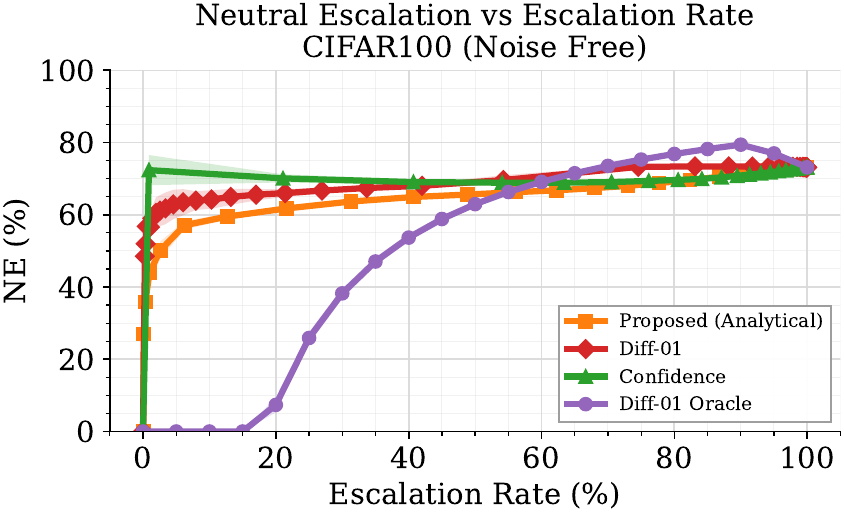}

\vspace{0.3em}

\includegraphics[width=\linewidth]{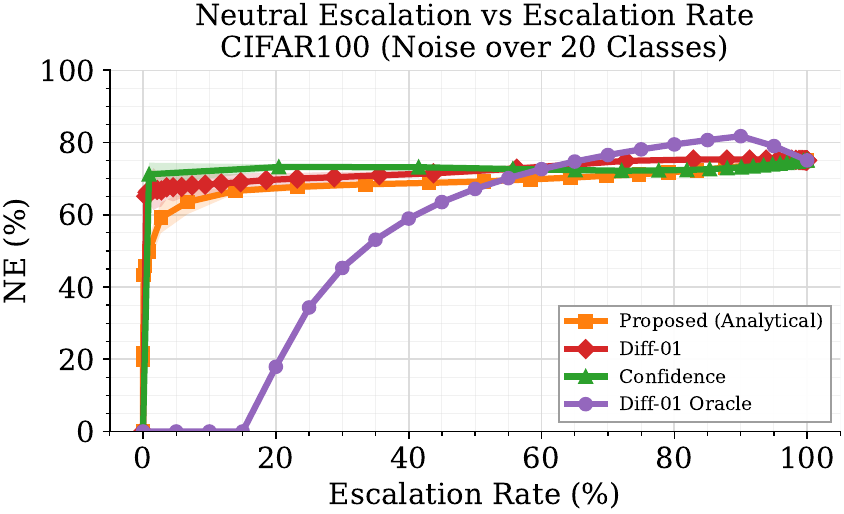}

\end{minipage}
\hfill
\begin{minipage}[t]{0.325\textwidth}
\centering

\includegraphics[width=\linewidth]{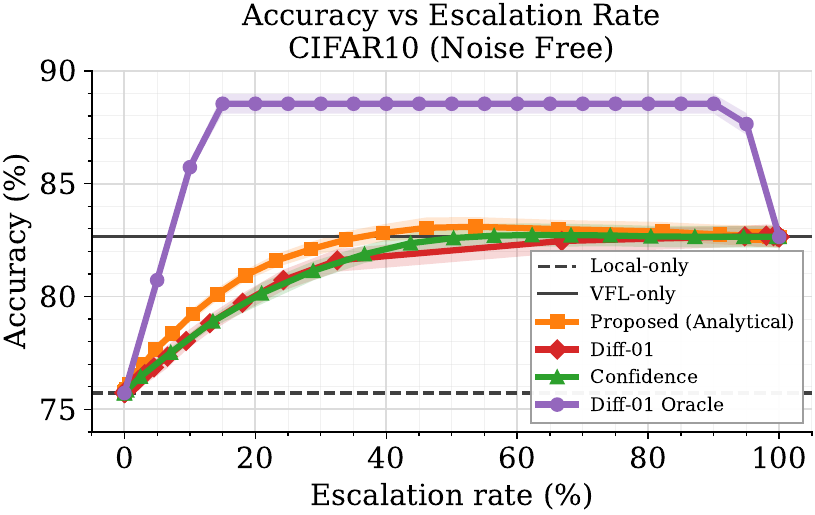}

\vspace{0.3em}

\includegraphics[width=\linewidth]{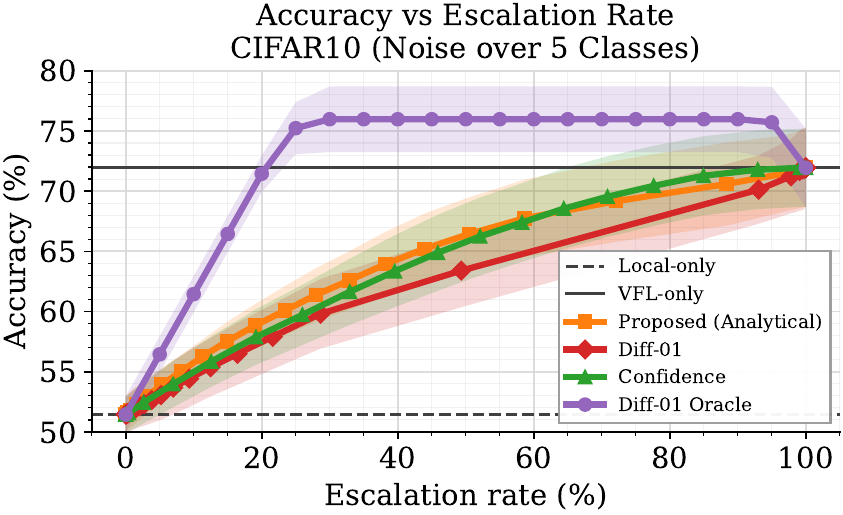}

\end{minipage}
\hfill
\begin{minipage}[t]{0.325\textwidth}
\centering

\includegraphics[width=\linewidth]{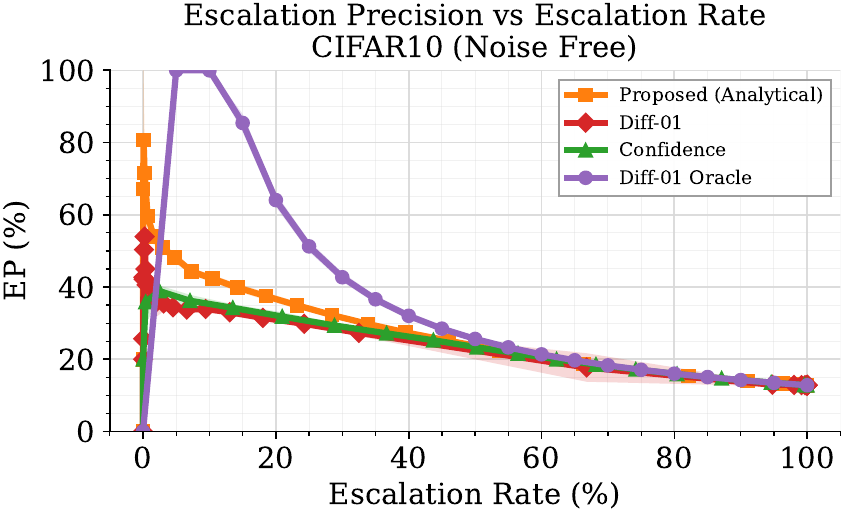}

\vspace{0.3em}

\includegraphics[width=\linewidth]{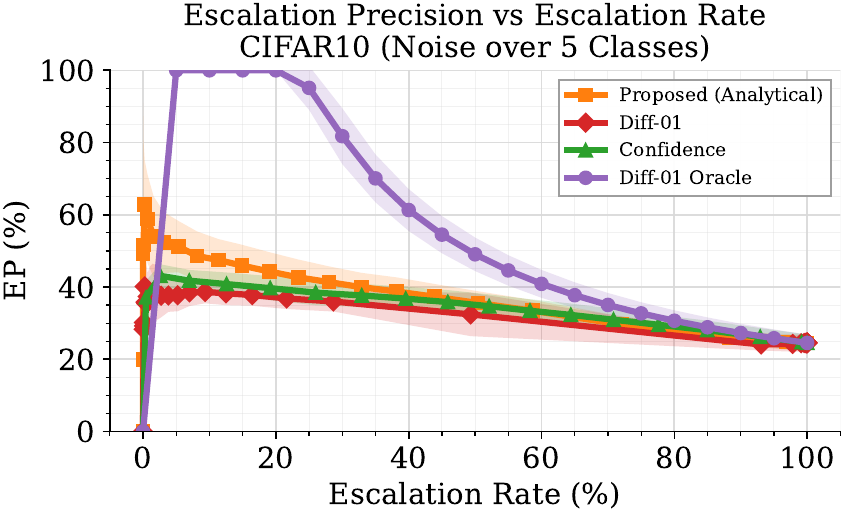}
\end{minipage}
\hfill
\begin{minipage}[t]{0.325\textwidth}
\centering

\includegraphics[width=\linewidth]{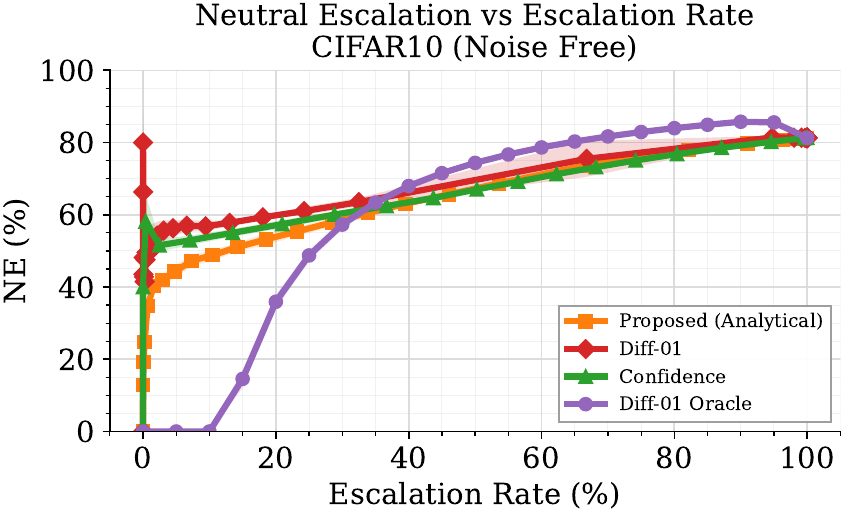}

\vspace{0.3em}

\includegraphics[width=\linewidth]{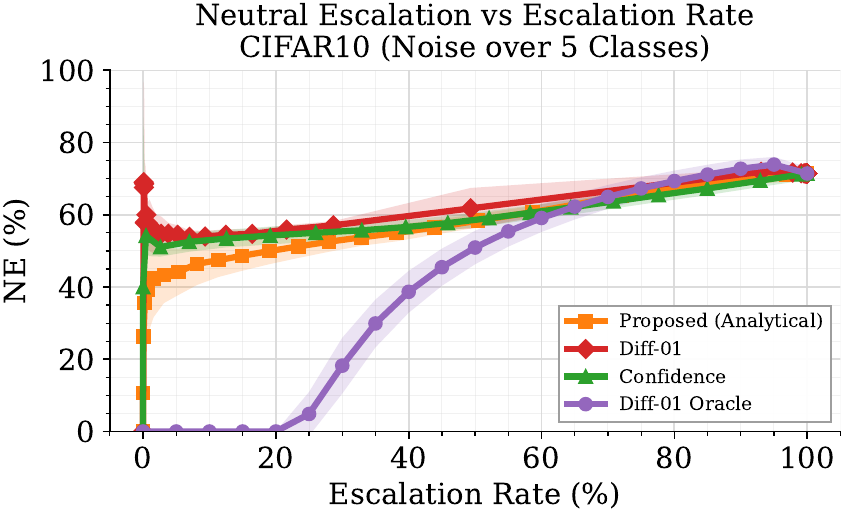}

\end{minipage}
\caption{Comparison of the proposed analytical expected-gain router, the learned router, the oracle, and a confidence-based baseline on the CIFAR-100 and CIFAR-10 datasets. For CIFAR-100, results are reported for the no-noise setting and for noise applied to 20 classes; for CIFAR-10, results are reported for the no-noise setting and for noise applied to 5 classes. Columns report accuracy, escalation precision (EP), and the proportion of neutral escalations (NE), each shown as a function of the escalation rate.}
\label{fig:cifar}
\end{figure*}

Finally, \autoref{fig:cifar} shows analogous behavior on CIFAR-10 and CIFAR-100 in the noise-free setting and under inference-time noise. Most examples are again neutral with respect to escalation, so aggressive thresholds over-escalate without commensurate accuracy gains. Consistent with the ModelNet40 findings, the analytical router provides the most favorable trade-off across operating points, improving over both endpoint policies and outperforming confidence-based  and learned-gain routing, while still leaving a gap to the oracle that highlights remaining headroom.

\section{Limitations}
\label{sec:limitations}

The proposed router is intentionally simple and interpretable, but it has several limitations. First, the analytical score relies on a classwise reliability approximation for the Round--2 VFL model. In particular, it replaces the sample-dependent quantity \(P(\hat y_{\mathrm{VFL}}=k \mid y=k,\bar{\boldsymbol p})\) with the class-level estimate \(P(\hat y_{\mathrm{VFL}}=k \mid y=k)\). This makes the score easy to estimate from held-out data, but it ignores within-class variation in the behavior of the collaborative model. If some samples from the same class benefit strongly from escalation while others do not, the classwise approximation may overestimate or underestimate the true gain.

Second, the method depends on the quality and representativeness of the calibration set. The calibrated posterior and the Round--2 reliability coefficients are both estimated offline. If the calibration distribution differs from the test distribution, the expected-gain score may become less reliable as observed in the simulations once noise is injected to the client views during inference. This is especially relevant under distribution shift or when some classes are rare, since the corresponding reliability estimates may have high variance.

\section{Conclusion}

We considered inference-time selective escalation in a two-round collaborative inference setting, where a cheap Round--1 local route prediction can optionally be replaced by a stronger but more expensive collaborative prediction. We formulated this routing problem as an expected-gain decision under a communication budget and proposed a simple analytical routing rule based on post-hoc calibration of a Round--1 posterior together with reliability estimates of the Round--2 route predictor. The resulting method is lightweight, interpretable, and does not require training a dedicated routing model. Empirically, the proposed rule improves  the communication--accuracy trade-off over confidence-based, learned-gain, and learning-to-defer baselines across the studied benchmarks.
\bibliographystyle{IEEEtran}
\bibliography{refs}

@article{pmlr-v70-guo17a,
  author       = {Chuan Guo and
                  Geoff Pleiss and
                  Yu Sun and
                  Kilian Q. Weinberger},
  title        = {On Calibration of Modern Neural Networks},
  journal      = {CoRR},
  volume       = {abs/1706.04599},
  year         = {2017},
  url          = {http://arxiv.org/abs/1706.04599},
  eprinttype   = {arXiv},
  eprint       = {1706.04599},
  bibsource    = {dblp computer science bibliography, https://dblp.org}
}

@inproceedings{gao2011active,
author = {Gao, Tianshi and Koller, Daphne},
title = {Active classification based on value of classifier},
year = {2011},
isbn = {9781618395993},
publisher = {Curran Associates Inc.},
address = {Red Hook, NY, USA},
booktitle = {Proceedings of the 25th International Conference on Neural Information Processing Systems},
pages = {1062–1070},
numpages = {9},
location = {Granada, Spain},
series = {NIPS'11}
}

@article{mozannar2020consistent,
  author       = {Hussein Mozannar and
                  David A. Sontag},
  title        = {Consistent Estimators for Learning to Defer to an Expert},
  journal      = {CoRR},
  volume       = {abs/2006.01862},
  year         = {2020},
  url          = {https://arxiv.org/abs/2006.01862},
  eprinttype   = {arXiv},
  eprint       = {2006.01862},
  bibsource    = {dblp computer science bibliography, https://dblp.org}
}

@article{flpaper,
  author       = {Peter Kairouz and
                  H. Brendan McMahan and
                  Brendan Avent and
                  Aur{\'{e}}lien Bellet and
                  Mehdi Bennis and
                  Arjun Nitin Bhagoji and
                  Kallista A. Bonawitz and
                  Zachary Charles and
                  Graham Cormode and
                  Rachel Cummings and
                  Rafael G. L. D'Oliveira and
                  Salim El Rouayheb and
                  David Evans and
                  Josh Gardner and
                  Zachary Garrett and
                  Adri{\`{a}} Gasc{\'{o}}n and
                  Badih Ghazi and
                  Phillip B. Gibbons and
                  Marco Gruteser and
                  Za{\"{\i}}d Harchaoui and
                  Chaoyang He and
                  Lie He and
                  Zhouyuan Huo and
                  Ben Hutchinson and
                  Justin Hsu and
                  Martin Jaggi and
                  Tara Javidi and
                  Gauri Joshi and
                  Mikhail Khodak and
                  Jakub Kone{\v{c}}n{\'y} and
                  Aleksandra Korolova and
                  Farinaz Koushanfar and
                  Sanmi Koyejo and
                  Tancr{\`{e}}de Lepoint and
                  Yang Liu and
                  Prateek Mittal and
                  Mehryar Mohri and
                  Richard Nock and
                  Ayfer {\"{O}}zg{\"{u}}r and
                  Rasmus Pagh and
                  Mariana Raykova and
                  Hang Qi and
                  Daniel Ramage and
                  Ramesh Raskar and
                  Dawn Song and
                  Weikang Song and
                  Sebastian U. Stich and
                  Ziteng Sun and
                  Ananda Theertha Suresh and
                  Florian Tram{\`{e}}r and
                  Praneeth Vepakomma and
                  Jianyu Wang and
                  Li Xiong and
                  Zheng Xu and
                  Qiang Yang and
                  Felix X. Yu and
                  Han Yu and
                  Sen Zhao},
  title        = {Advances and Open Problems in Federated Learning},
  journal      = {CoRR},
  volume       = {abs/1912.04977},
  year         = {2019},
  url          = {http://arxiv.org/abs/1912.04977},
  eprinttype   = {arXiv},
  eprint       = {1912.04977},
  bibsource    = {dblp computer science bibliography, https://dblp.org}
}

@misc{jitkrittum2023cascade,
      title={When Does Confidence-Based Cascade Deferral Suffice?}, 
      author={Wittawat Jitkrittum and Neha Gupta and Aditya Krishna Menon and Harikrishna Narasimhan and Ankit Singh Rawat and Sanjiv Kumar},
      year={2024},
      eprint={2307.02764},
      archivePrefix={arXiv},
      primaryClass={cs.LG},
      url={https://arxiv.org/abs/2307.02764}, 
}

@inproceedings{cifar,
  title={Learning Multiple Layers of Features from Tiny Images},
  author={Alex Krizhevsky},
  year={2009},
  url={https://api.semanticscholar.org/CorpusID:18268744}
}

@inproceedings{
mao2023twostage,
title={Two-Stage Learning to Defer with Multiple Experts},
author={Anqi Mao and Christopher Mohri and Mehryar Mohri and Yutao Zhong},
booktitle={Thirty-seventh Conference on Neural Information Processing Systems},
year={2023},
url={https://openreview.net/forum?id=GIlsH0T4b2}
}

@InProceedings{liu2024underfitting,
  title = 	 {Mitigating Underfitting in Learning to Defer with Consistent Losses},
  author =       {Liu, Shuqi and Cao, Yuzhou and Zhang, Qiaozhen and Feng, Lei and An, Bo},
  booktitle = 	 {Proceedings of The 27th International Conference on Artificial Intelligence and Statistics},
  pages = 	 {4816--4824},
  year = 	 {2024},
  editor = 	 {Dasgupta, Sanjoy and Mandt, Stephan and Li, Yingzhen},
  volume = 	 {238},
  series = 	 {Proceedings of Machine Learning Research},
  month = 	 {02--04 May},
  publisher =    {PMLR},
  url = 	 {https://proceedings.mlr.press/v238/liu24h.html}
}

@Article{khan2022commvfl,
AUTHOR = {Khan, Afsana and ten Thij, Marijn and Wilbik, Anna},
TITLE = {Communication-Efficient Vertical Federated Learning},
JOURNAL = {Algorithms},
VOLUME = {15},
YEAR = {2022},
NUMBER = {8},
ARTICLE-NUMBER = {273},
URL = {https://www.mdpi.com/1999-4893/15/8/273},
ISSN = {1999-4893},
DOI = {10.3390/a15080273}
}

@article{calibration_d,
  author       = {Meelis Kull and
                  Miquel Perell{\'{o}}{-}Nieto and
                  Markus K{\"{a}}ngsepp and
                  Telmo de Menezes e Silva Filho and
                  Hao Song and
                  Peter A. Flach},
  title        = {Beyond temperature scaling: Obtaining well-calibrated multiclass probabilities
                  with Dirichlet calibration},
  journal      = {CoRR},
  volume       = {abs/1910.12656},
  year         = {2019},
  url          = {http://arxiv.org/abs/1910.12656},
  eprinttype   = {arXiv},
  eprint       = {1910.12656},
  bibsource    = {dblp computer science bibliography, https://dblp.org}
}

@misc{khan2025vflreview,
      title={Vertical Federated Learning: A Structured Literature Review}, 
      author={Afsana Khan and Marijn ten Thij and Anna Wilbik},
      year={2023},
      eprint={2212.00622},
      archivePrefix={arXiv},
      primaryClass={cs.LG},
      url={https://arxiv.org/abs/2212.00622}, 
}

@misc{castiglia2023lessvfl,
      title={LESS-VFL: Communication-Efficient Feature Selection for Vertical Federated Learning}, 
      author={Timothy Castiglia and Yi Zhou and Shiqiang Wang and Swanand Kadhe and Nathalie Baracaldo and Stacy Patterson},
      year={2023},
      eprint={2305.02219},
      archivePrefix={arXiv},
      primaryClass={cs.LG},
      url={https://arxiv.org/abs/2305.02219}, 
}

@nonarchival{
inoue2023sparsevfl,
title={Sparse{VFL}: Communication-Efficient Vertical Federated Learning Based on Sparsification of Embeddings and Gradients},
author={Yoshitaka Inoue and Hiroki Moriya and Qiong Zhang and Kris Skrinak},
booktitle={International Workshop on Federated Learning for Distributed Data Mining},
year={2023},
url={https://openreview.net/forum?id=BVH3-XCRoN3}
}

@misc{sun2023oneshotvfl,
      title={Communication-Efficient Vertical Federated Learning with Limited Overlapping Samples}, 
      author={Jingwei Sun and Ziyue Xu and Dong Yang and Vishwesh Nath and Wenqi Li and Can Zhao and Daguang Xu and Yiran Chen and Holger R. Roth},
      year={2023},
      eprint={2303.16270},
      archivePrefix={arXiv},
      primaryClass={cs.LG},
      url={https://arxiv.org/abs/2303.16270}, 
}

@article{zhou2025vflcafe,
author = {Zhou, Jiahui and Liang, Han and Wu, Tian and Zhang, Xiaoxi and Yu, Jiang and Tan, Chee},
year = {2025},
month = {01},
pages = {66},
title = {VFL-Cafe: Communication-Efficient Vertical Federated Learning via Dynamic Caching and Feature Selection},
volume = {27},
journal = {Entropy},
doi = {10.3390/e27010066}
}

@article{franc2023reject,
  author  = {Vojtech Franc and Daniel Prusa and Vaclav Voracek},
  title   = {Optimal Strategies for Reject Option Classifiers},
  journal = {Journal of Machine Learning Research},
  year    = {2023},
  volume  = {24},
  number  = {11},
  pages   = {1--49},
  url     = {http://jmlr.org/papers/v24/21-0048.html}
}

@misc{hendrickx2024survey,
      title={Machine Learning with a Reject Option: A survey}, 
      author={Kilian Hendrickx and Lorenzo Perini and Dries Van der Plas and Wannes Meert and Jesse Davis},
      year={2024},
      eprint={2107.11277},
      archivePrefix={arXiv},
      primaryClass={cs.LG},
      url={https://arxiv.org/abs/2107.11277}, 
}

@misc{madras2018defer,
      title={Predict Responsibly: Improving Fairness and Accuracy by Learning to Defer}, 
      author={David Madras and Toniann Pitassi and Richard Zemel},
      year={2018},
      eprint={1711.06664},
      archivePrefix={arXiv},
      primaryClass={stat.ML},
      url={https://arxiv.org/abs/1711.06664}, 
}

@article{resnet18,
  author       = {Kaiming He and
                  Xiangyu Zhang and
                  Shaoqing Ren and
                  Jian Sun},
  title        = {Deep Residual Learning for Image Recognition},
  journal      = {CoRR},
  volume       = {abs/1512.03385},
  year         = {2015},
  url          = {http://arxiv.org/abs/1512.03385},
  eprinttype    = {arXiv},
  eprint       = {1512.03385},
  bibsource    = {dblp computer science bibliography, https://dblp.org}
}

\section*{Biography}
\vspace{-113pt}
\begin{IEEEbiographynophoto}{Mohamad Mestoukirdi}
(Member, IEEE) was born in Tyre, Lebanon, in 1995. He received a double degree in engineering from the \mbox{Politecnico di Torino} (Polito), Turin, Italy, and the Lebanese University, Beirut, Lebanon, in 2019, and the Ph.D. degree from Sorbonne University, Paris, France, in 2023. Since 2023, he has been a Researcher with the Synergistic Autonomous Systems Group (formerly the Wireless Communication Systems Group) at Mitsubishi Electric R\&D Centre Europe, Rennes, France.
\end{IEEEbiographynophoto}
\vspace{-113pt}
\begin{IEEEbiographynophoto}{Vincent Corlay}
(Member, IEEE) was born in Rennes, France, in 1993. He received the Engineering degree from the National Institute of Applied Sciences (INSA), Rennes, France, in 2017 and the Ph.D. degree from the Institut Polytechnique de Paris, Télécom Paris, in 2020. Since 2020, he has been a permanent Researcher with the group Synergystic Autonomous Sytems (formely Wireless Communication Systems), Mitsubishi Electric R\&D Centre Europe, Rennes. In 2021, he received the Best Thesis Award Runner-Up Award from the Institut Polytechnique de Paris.\end{IEEEbiographynophoto}

\end{document}